\pdfoutput=1
\documentclass[11pt]{article}

\usepackage{enumitem}
\usepackage{amsmath}
\usepackage{amsfonts}
\usepackage{tabularx}
\usepackage{array}
\usepackage{makecell}
\usepackage{booktabs}
\usepackage{caption}
\usepackage{soul}
\usepackage{xcolor}
\usepackage{amssymb}
\usepackage{multirow}
\usepackage{graphicx}
\usepackage{placeins}
\usepackage{dsfont}
\usepackage{float}

\usepackage{xurl}

\definecolor{lightblue}{RGB}{212,226,255}
\definecolor{lightorange}{RGB}{255,228,168}
\newcommand{\tighthlblue}[1]{{\sethlcolor{lightblue}\hl{#1}}}
\newcommand{\tighthlorange}[1]{{\sethlcolor{lightorange}\hl{#1}}}

\usepackage[final]{acl}

\usepackage{times}
\usepackage{latexsym}

\usepackage[T1]{fontenc}
\usepackage[utf8]{inputenc}

\usepackage{microtype}

\usepackage{inconsolata}

\usepackage{graphicx}

\title{Counterfactual Simulatability of LLM Explanations for Generation Tasks}

\author{Marvin Limpijankit, 
    Yanda Chen,
    Melanie Subbiah, \\
    \textbf{Nicholas Deas}, and
    \textbf{Kathleen McKeown} \\
    Department of Computer Science, Columbia University \\
    \texttt{ml4431@columbia.edu}}

\begin{document}
\maketitle
\begin{abstract}

LLMs can be unpredictable, as even slight alterations to the prompt can cause the output to change in unexpected ways. Thus, the ability of models to accurately explain their behavior is critical, especially in high-stakes settings. Counterfactual simulatability measures how well an explanation allows users to infer the model's output on related counterfactuals and has been previously studied for yes/no question answering. We provide a general framework for extending this method to generation tasks, using news summarization and medical suggestion as example use cases. We find that while LLM explanations do enable users to better predict their outputs on counterfactuals in the summarization setting, there is significant room for improvement for medical suggestion. Furthermore, our results suggest that evaluating counterfactual simulatability may be more appropriate for skill-based tasks as opposed to knowledge-based tasks.

\end{abstract}

\section{Introduction}

While large language models (LLMs) have proven effective for a diverse range of applications, their outputs still often contain hallucinations of unsupported information \citep{ji2023survey} or biases \citep{sheng-etal-2021-societal} that hinder their reliability on critical language generation tasks. At the same time, it is important that users themselves can accurately evaluate the model capabilities, knowledge, and reliability \citep{steyvers-think}. Particularly in high stakes domains, such as medical applications, misunderstandings or the lack of ability to predict model behavior on unseen inputs can pose disastrous risks to users \citep{michalowski-explain}.

To anticipate these risks, recent work has turned attention to evaluating the reliability of LLM explanations \citep{madsen-etal-2024-self, turpin-cot, kunz-kuhlmann-2024-properties}. In particular, \citet{chen-simulatbility} evaluates the \textit{counterfactual simulatability} of LLM explanations for yes/no question answering. Counterfactual simulatability is a measure of how well a model's explanation allows humans to correctly infer the model’s predictions on simulatable counterfactuals (unseen inputs where the explanation should enable the user to confidently guess the model's output). Furthermore, according to \citet{chen-simulatbility}, the counterfactual simulatability of an explanation can be decomposed into \textit{simulation generality}, a measure of the diversity of simulatable counterfactuals and \textit{simulation precision}, a measure of the proportion of these counterfactuals for which humans correctly infer the model’s output. Ideal model explanations should balance both generality and precision. 

\begin{figure*}[h]
\centering
\includegraphics[width=\textwidth]{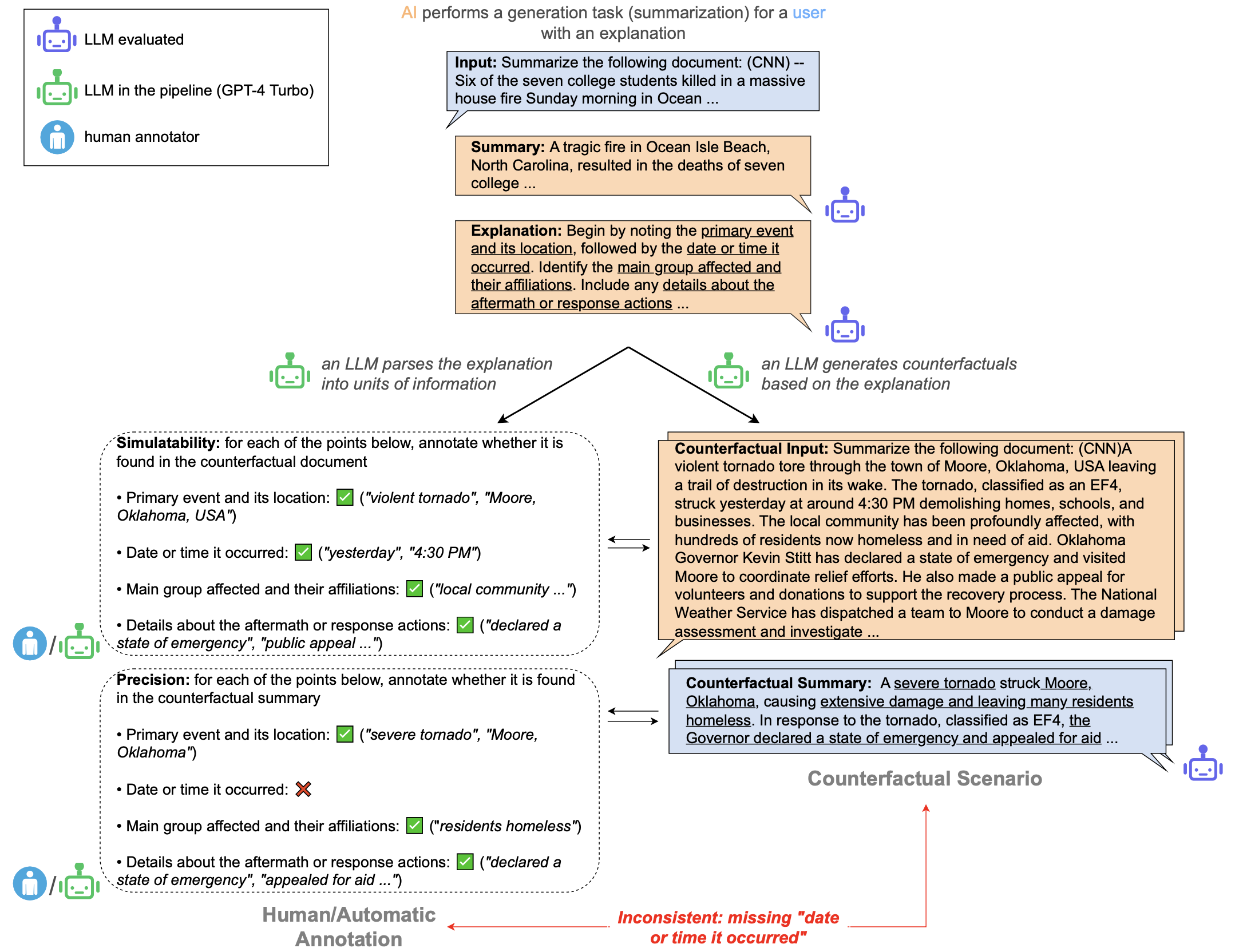}
\caption{Our evaluation pipeline. Given a model's explanation, an LLM is prompted to decompose the explanation into atomic units (left) and generate relevant counterfactuals (right). For each unit, an annotator verifies whether the element appears in the counterfactual (simulatability) and the counterfactual output (precision).}
\label{fig:evaluation_pipeline}
\end{figure*}

Counterfactual simulatability, however, is equally critical to generation tasks, where the larger space of possible outputs makes understanding a model’s decision process more challenging. To fill this gap, we formalize a framework for evaluating counterfactual simulatability in language generation tasks (Figure \ref{fig:evaluation_pipeline}). To evaluate a model's explanation, we first use a separate LLM to (1) decompose the explanation into atomic units and (2) generate relevant counterfactuals. Then, an annotator (human or the LLM) evaluates the simulatability and precision of each unit of the model’s explanation using the counterfactual and the model’s output on the counterfactual respectively. Finally, generality and precision scores are calculated following \S\ref{section:estimating_mental_models}. \textit{Rather than evaluating the factual correctness of explanations, our framework measures the ability of LLMs to accurately describe their behavior in a generative setting. We evaluate whether LLMs' explanations lead to reliable human mental models that are consistent with their outputs}.

We apply the framework to two tasks: news summarization using CNN/DM \citep{nallapati-etal-2016-abstractive} and medical suggestion generation using the Taiwan e-Hospital Dataset \citep{chen-etal-2022-learning-generate}. These tasks involve different \textit{trade-offs between generality and precision}. Explanations for news summarization can be highly general as models may employ similar approaches for documents that contain common elements (e.g., dates, quotes) but differ in content. In contrast, medical suggestion explanations may be less general as the model's response can be highly dependent on the user input (i.e., minor variances in expressed symptoms can lead to different suggestions). This can lead to a more limited set of simulatable counterfactuals, though potentially being more precise in its outputs. With these generation tasks, we conduct an initial evaluation of LLMs’ explanations and assess where models’ explanations in these generation tasks fail. 

In summary, our contributions are as follows:
\begin{enumerate}[itemsep=5pt]
    \item We propose a framework to measure the ability of LLMs to accurately describe their behavior in a generative setting by evaluating whether LLMs' explanations lead to mental models consistent with their outputs (counterfactual simulatability).
    \item We assess the feasibility of using an LLM to automate human annotation in our evaluation pipeline and find that doing so achieves agreement with humans comparable to agreement among human annotators.
    \item Using our framework, we evaluate Chain-of-thought and Post-hoc explanations for multiple LLMs on two complementary tasks: news summarization and medical suggestion generation. We show that LLM explanations lead to reliable mental models in news summarization, but not medical suggestion. \footnote{Our code is available at \url{github.com/mlimpijankit/counterfactual-simulatability-generation-tasks}}
\end{enumerate}

\section{Related Work}

\textbf{Human mental models. }
Humans form mental models of the physical world as a whole \citep{gentner-mental} as well as specific technologies (e.g., \citealp{payne-mental,du-voice,lei-mobile}) through their past experiences and observations. Specifically with regard to artificial intelligence systems, explanations of model predictions have been considered high quality if they provide users with an accurate and generalizable understanding of the system in the form of mental models \citep{rutjes-xai, merry-xai}. When such explanations are successful, they can improve users' ability to effectively use AI models \citep{vasconcelos-cost, senoner-collab} as well as to anticipate and correct undesirable model behaviors, such as biases and incorrect predictions \citep{bansal-accuracy}. We specifically evaluate LLMs' explanations in generation tasks considering their ability to help form mental models.

\textbf{Explanation evaluation. }
A variety of different approaches have been used to evaluate the quality and utility of model-generated natural language explanations and rationales. Studies evaluating natural language explanations similar to word attributions \citep{huang-explain, madsen-etal-2024-self} as well as unconstrained explanations \citep{turpin-cot} have focused on faithfulness measures. In particular, work has proposed metrics for dimensions including comprehensiveness \citep{deyoung-eraser}, sufficiency \citep{deyoung-eraser}, alignment with human rationales \citep{fayyaz-alignment}, plausibility \citep{wojciechowski2024faithful}, and scrutability \citep{zhichao-scrutability} among others. In contrast to intrinsic measures of model explanations, other work evaluates explanations extrinsically through impact on task performance (e.g., \citealp{camburu-esnli,wei-cot,krishna-posthoc}). Among these topics, relatively few works have investigated natural language explanations in language generation tasks; such studies include dialogue responses \citep{zhou-commonsense}, dialogue understanding \citep{gao-dialogue}, and more prominently, open-ended question answering \citep{ho-wikiwhy,fragkathoulas-qa,lyu-faithful}. While \citet{chen-simulatbility} introduces \textit{counterfactual simulatability} in a binary classification setting, we propose a novel framework for counterfactual simulatability in language generation settings.  

\section{Counterfactual Simulatability for Generation Tasks}

Explanations can be evaluated by considering whether an observer, having seen a model's explanation for some input, can infer (i.e., simulate) the model's output on a related counterfactual input. For instance, if a user asks ``\textit{do dolphins swim?}'' and a model answers ``\textit{yes}'' with the explanation ``\textit{all aquatic animals swim}'', then the user would infer that when asked the counterfactual ``\textit{do starfish swim?}'', the model will similarly answer ``\textit{yes}''. If, in reality,  the model answers ``\textit{no}'' to this question, then, as \citet{chen-simulatbility} notes, the explanation is ineffective because it creates a mental model that is inconsistent with the model's behavior, even though the answer may be factually correct. More specifically, counterfactual simulatability measures how accurate these mental models are on simulatable counterfactuals, unseen inputs where the explanation allows the user to confidently predict the model's output \citep{chen-simulatbility}.

However, in contrast to the classification example, generation tasks have a much larger output space (e.g., there exists many possible summaries for a news document). This makes simulation extremely challenging as it is impossible to precisely identify a single possible output based on an explanation. For instance in news summarization, if a user is shown the explanation ``\textit{the summary should include the key event}'' and a counterfactual document on ``\textit{the opening ceremony of the 2024 Paris Olympics}'', while the user can logically infer that the summary will include the opening ceremony, it is impossible to predict the exact wording in which it will appear. Explanations typically cannot enable humans to pinpoint a single model output. However, they are still very useful if they help humans narrow down the possible outputs (e.g., refining ``all possible summaries'' to ``summaries that mention the opening ceremony'').

\subsection{Notation}

For a given generation task, a model $M$ takes an input $x \in X$ and produces an output $o_x \in O$ and a corresponding explanation $e_x$. Here, the input, output, and explanation are all natural language and, in the case of generation, $|O|$ may be arbitrarily large. A human observes $x, e_x$, and forms a one-to-many mental model $h_{x, e_x}: X \rightarrow \mathcal{P}(O)$, where $\mathcal{P}(O)$ denotes the power set of $O$, and $h_{x, e_x}(x')$ denotes what the human infers to be $M$'s possible outputs on a counterfactual $x'$. For simplicity, $h_{e_x}(x')$ is used to denote $h_{x, e_x}(x')$.

\begin{figure*}[h]
\centering
\includegraphics[width=0.97\textwidth]{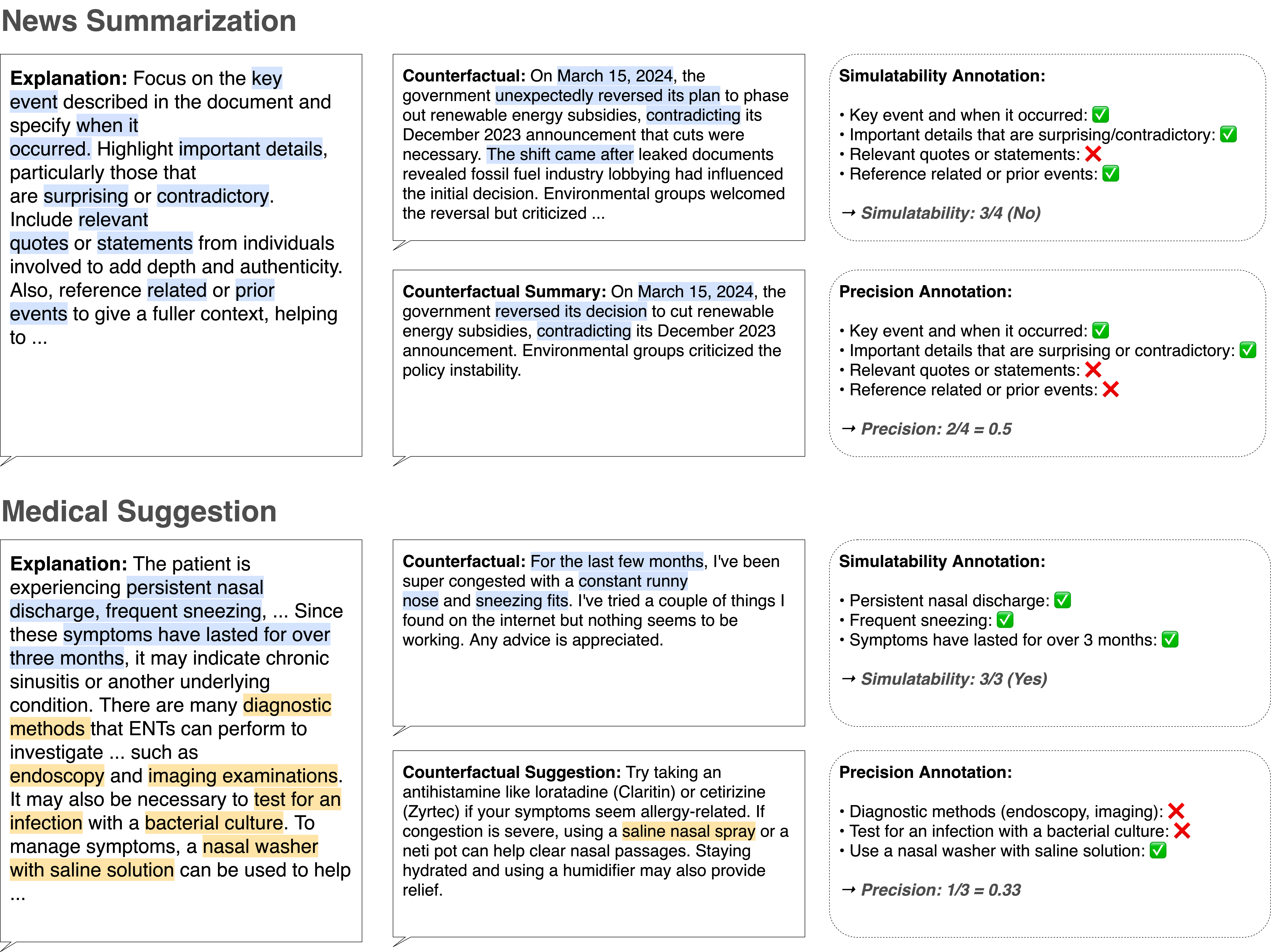}
\caption{Example explanations, counterfactuals, counterfactual outputs, and annotations for news summarization and medical suggestion. Atomic units of the explanation are highlighted (for medical suggestion, \tighthlblue{blue: patient information}, \tighthlorange{orange: suggestions}).}
\label{fig:annotation_examples}
\end{figure*}

\subsection{Simulatability}

Simulatability is a metric on counterfactuals that describes whether the model's output on that counterfactual can be inferred given an explanation. More specifically, a simulatable counterfactual is one where an observer, using the mental model produced by an explanation, can refine their expectation of the model's output (i.e., $ | h_{x, e_x}(x') | \ll O $). Non-simulatable counterfactuals are omitted from our evaluation as the explanations are not helpful in these cases.

\subsection{Simulation generality}
\label{section:class_generality}
Simulation generality is a metric on explanations that measures the diversity of simulatable counterfactuals an explanation leads to. For instance, the explanation ``\textit{the summary should include the key event}'' is more general compared to ``\textit{the summary should mention the opening ceremony of the 2024 Paris Olympics}'' since the former can be applied to documents covering a broader range of topics. It is calculated as one minus the average pairwise similarity between simulatable counterfactuals:

\begin{equation}
    \text{generality} = 1 - \mathbb{E}_{x', x'' \sim p, x' \neq x''}[\alpha(x', x'')]
\label{eq:generality}
\end{equation}

\noindent where $p$ is the distribution of simulatable counterfactuals and $\alpha$ is a similarity metric such as cosine similarity. Explanations with higher generality are more useful since they enable users to infer the model's behavior on a wider range of possible inputs. 

\subsection{Simulation precision}
\label{section:class_precision}
Simulation precision is another metric on explanations and measures whether the model's actual output $M(x')$ is within the observer's inferred output space: 

\begin{equation}
    \text{precision} = \mathbb{E}_{x' \sim p} \left [ \mathbf{1} [M({x'}) \in h_{e_x}(x')] \right ]
\end{equation}

\noindent Explanations with higher precision scores indicate that the mental models they produce are more aligned with the model's actual behavior. In practice, it is unfeasible for humans to enumerate all valid outputs in $h_{e_x}(x')$ due to the large output space. Thus, in \S \ref{section:estimating_mental_models} we propose a method to estimate mental models by evaluating the atomic units of a model’s explanation.

\section{Methods}

\subsection{Datasets}

We select two generative tasks that represent practical LLM use cases and whose natural language explanations differ in generality and precision. For the first task of \textbf{news summarization}, we use the CNN/DM dataset \citep{nallapati-etal-2016-abstractive}. Summarization is chosen as it is a well-studied application for language generation. Additionally, this task allows us to study high-level, abstract explanations since summarization does not rely on facts learned during the LLM's pre-training, rather its ability to extract key elements from the input document.

Conversely, we also investigate \textbf{medical suggestion}, a domain where explanations are critical. For this task, we use the Taiwan e-Hospital Dataset, a collection of 86,399 Mandarin question answer pairs from the online health website Taiwan e-Hospital \citep{chen-etal-2022-learning-generate}. The data was translated into English using the Google Translate API. We select this specific dataset because each sample consists of a question, suggestion, explanation triplet, ensuring that the questions are complex enough such that explanations are helpful. Furthermore, the dataset also mirrors the practical setting we aim to study, where everyday users interact with LLMs to seek general medical advice without any specialized knowledge. 

The contrast between explanations for these tasks is illustrated in Figure \ref{fig:annotation_examples}. Medical suggestions are more knowledge-based, requiring the LLM to identify key aspects of the query (e.g. expressed symptoms, relevant medical history) and generate suggestions using knowledge encoded in the model. As such, the explanations are highly specific to the input content. In the example above, the summarization explanation applies to counterfactuals containing any ``\textit{key event}'' or ``\textit{quotes or statements}' whereas the medical suggestion explanation can only be used to infer the model's output when a user expresses ``\textit{persistent nasal discharge}'', ``\textit{frequent sneezing}'', and ``\textit{symptoms [that] have lasted for over three months}''. While medical suggestion explanations likely apply to a more limited set of counterfactuals, they allow the user to infer specific pieces of information rather than high-level elements as is the case for summarization. Therefore, these tasks are complementary for studying counterfactual simulatability as they represent different requirements for generality and precision.

\subsection{Models}

Our proposed framework uses two models: the LLM being evaluated and the LLM used to automate parts of the pipeline (e.g., counterfactual generation, explanation decomposition, annotation). We evaluate Anthropic’s Claude 3.7 Sonnet, OpenAI’s GPT-4 and GPT-4 Turbo, and Meta’s Llama 3.3 70B Instruct. These represent a few popular proprietary and open-source models. To automate steps in the pipeline, we opt for GPT-4 Turbo due to its competitive performance at a low cost.

\subsection{Explanations}
\label{section:explanations}

Through prompting (instructions, few-shot examples), we guide the LLMs to generate explanations that emphasize the difference in generality and precision across tasks. For news summarization, to be highly generalizable, we encourage the model to explain its decision process at a high-level, identifying abstract elements while avoiding reference to specific topics. For medical suggestion, we instruct the model to first identify important details in the user’s question, propose possible underlying causes, and suggest recommended actions accordingly. These approaches reflect the skill-based vs. knowledge-based nature of our tasks. Additionally, following \citet{chen-simulatbility}, we experiment with both Chain-of-thought and Post-hoc prompting to produce explanations \citep{camburu-esnli}. The prompts used are provided in Appendix \ref{app:llm_prompts}.

\subsection{Estimating mental models}
\label{section:estimating_mental_models}

Generation tasks are challenging for counterfactual simulatability because mental models are extremely difficult to estimate given the large space of possible outputs. Motivated by previous work on atomic units \citep{wright2022generating, chen2022generating, kamoi2023wice}, we introduce a method for estimating mental models by decomposing explanations into units of information as a proxy.

Intuitively, explanations $e_x$ identify key pieces of information that the model considers essential for the output. Thus, an observer may simulate that if the counterfactual input $x'$ includes a piece of information $a$ deemed important in $e_x$, the model should similarly take into account $a$ in its output for $x'$. Based on this explanation $e_x$, an observer's mental model $h_{e_x}$ can be formalized as: 

\begin{equation}
    h_{e_x}(x') = \{o \in O \mid \forall a \in e_x \cap x', a \in o\}
\end{equation}

\noindent where the space of possible outputs on the counterfactual is refined to only outputs that contain the elements $\{a\}$. With this, we formulate simulatability, generality, and precision for each explanation, counterfactual pair as follows. First, the simulatability of a counterfactual is determined by verifying if all atomic units of the explanation appear in the counterfactual:

\begin{equation}
    % \text{simulatability} = \frac{|\{ a \in e_x \cap x'\}|}{| \{ a \in e_x \} |}.
    \text{simulatability} = \mathds{1} \{\forall a \in e_x, a \in x'\}
\label{eq:simulatability}
\end{equation}

Then, all non-siumulatable counterfactuals are discarded and the generality of an explanation is calculated as one minus the average pairwise cosine similarity between simulatable counterfactuals (Equation \ref{eq:generality}). Finally, by identifying whether each unit of the explanation is addressed in the model's output on the counterfactual, a precision score is calculated as: 

\begin{equation}
    \text{precision} = \frac{|\{ a \in e_x \cap M(x')\}|}{|\{ a \in e_x\}|}
\label{eq:precision}
\end{equation}

\noindent Note that this differs from the formulation in \S \ref{section:class_precision}. Precision scores are averaged across counterfactuals for a given explanation. 

\paragraph{Explanation Decomposition.}

The atomic units of an explanation should highlight aspects that are considered important to the model, and thus, a user would expect in the output accordingly. These effectively describe a user's mental model and should ideally be separate, non-overlapping points. One key distinction between the tasks is that while each atomic unit of a summarization explanation is linked to an expected item in the input (for simulatability) and its reference in the output (for precision), medical suggestion explanations do not follow this one-to-one mapping of units between input and output. For example in Figure \ref{fig:annotation_examples}, while ''\textit{Persistent nasal discharge}'' is identified as important, it is unclear which suggested action the model has generated in response to this symptom. To address this, for the medical domain, we further instruct the LLM when decomposing the explanation to classify units into two categories, patient information and suggestions. Atomic units from these categories are then used to evaluate simulatability and precision respectively during annotation (Figure \ref{fig:annotation_examples}).

\subsection{Experimental setup}
\label{section:experimental_setup}

We run two main experiments: (1) a human evaluation where GPT-4 Turbo explanations are evaluated using human annotations and (2) a larger-scale, fully automatic evaluation of multiple models (Claude 3.7 Sonnet, GPT-4, Llama 3) using GPT-4 Turbo as the annotator. This allows us to assess the feasibility of using an LLM as an annotator before adopting the automatic approach that can scale to larger experiments.

\paragraph{Human evaluation.} Using randomly sampled inputs from our dataset, we prompt GPT-4 Turbo to \textit{generate outputs and explanations}. Note that here, GPT-4 Turbo is both the model being evaluated and the model being used in the pipeline. For each explanation, we use GPT-4 Turbo to \textit{generate 3 relevant counterfactuals}, providing the explanation in the context. Then, we use GPT-4 Turbo to \textit{decompose the explanation into atomic units} (i.e., $e_x \rightarrow \{a\}$) and, in the medical suggestion use case, instruct it to extract units into two groups for patient details or suggested actions. Finally, we \textit{generate counterfactual outputs} using the same prompt as the first step and have humans examine the counterfactual and counterfactual output to \textit{annotate simulatability and precision} for each atomic unit of the explanation. 

Additionally, we assess the ability of GPT-4 Turbo to break down explanations by instructing the annotators to indicate whether each unit was extracted correctly and whether details that should have been extracted are missing. We also briefly investigate if increasing the number of counterfactuals generated per explanation affects generality, but find no noticeable differences (Appendix \ref{app:cf_gen_generality}). 

\iffalse

\begin{table*}[h]
\centering
\setlength{\tabcolsep}{5pt}
\renewcommand{\arraystretch}{1.3}
\normalsize
\begin{tabular}{l|cccc|c||cccc|c}
\hline
& \multicolumn{5}{c||}{\textbf{News Summarization}} & \multicolumn{5}{c}{\textbf{Medical Suggestion}} \\
\hline
Annotator & 1 & 2 & 3 & \small{GPT-4 Turbo} & n & 1 & 2 & 3 & \small{GPT-4 Turbo} & n \\
\hline
1 & - & 0.35 & 0.71 & 0.64 & 263 & - & 0.76 & 0.74 & 0.68 & 273 \\
2 & 0.35 & - & 0.57 & 0.48 & 213 & 0.76 & - & 0.73 & 0.54 & 261 \\
3 & 0.71 & 0.57 & - & 0.71 & 73 & 0.74 & 0.73 & - & 0.74 & 258 \\
\hline
\end{tabular}
\caption{Inter-annotator agreement (Cohen's Kappa) between human annotators and GPT-4 Turbo for news summarization and medical suggestion.}
\label{tab:inter-annotator-agreement}
\end{table*}

\fi

\begin{table*}[h]
\centering
\setlength{\tabcolsep}{5pt}
\setlength{\arrayrulewidth}{0.8pt}
\renewcommand{\arraystretch}{1.3}
\normalsize
\begin{tabular}{|c|cccc|c||cccc|c|}
\hline
& \multicolumn{5}{c||}{\textbf{News Summarization}} & \multicolumn{5}{c|}{\textbf{Medical Suggestion}} \\
\hline
Annotator & 1 & 2 & 3 & \normalsize{GPT-4 Turbo} & n & 1 & 2 & 3 & \normalsize{GPT-4 Turbo} & n \\
\hline
1 & - & 0.35 & 0.71 & 0.64 & 263 & - & 0.76 & 0.74 & 0.68 & 273 \\
2 & 0.35 & - & 0.57 & 0.48 & 213 & 0.76 & - & 0.73 & 0.54 & 261 \\
3 & 0.71 & 0.57 & - & 0.71 & 73 & 0.74 & 0.73 & - & 0.74 & 258 \\
\hline
\end{tabular}
\caption{Inter-annotator agreement (Cohen's Kappa) between human annotators and GPT-4 Turbo for news summarization and medical suggestion.}
\label{tab:inter-annotator-agreement}
\end{table*}

We assigned 3 student annotators per task to evaluate 15 explanations along with 3 generated counterfactuals each (45 explanation, counterfactual pairs total). Each annotator was given an overlapping set of 3 explanations (to measure human-human agreement) plus an additional 4. Since they annotated the atomic units of each explanation, the exact number of annotations varied but approximately resulted in 260 annotations per person (see Table \ref{tab:human_eval_results}). The annotators were not required to have any specialized domain knowledge. One annotator (news summarization annotator \#3 in Table \ref{tab:inter-annotator-agreement}) did not complete all annotations, leading to a slight discrepancy in annotations across tasks. The prompts and annotations instructions are provided in Appendix \ref{app:llm_prompts} and \ref{app:human_eval} respectively.

\paragraph{Comparison of human-LLM annotations.} We evaluate the feasibility of using an LLM to automate human annotation by prompting GPT-4 Turbo with similar instructions and measuring the inter-annotator agreements between human-human and human-LLM pairs.

\paragraph{Automatic evaluation.} Finally, using GPT-4 Turbo as the annotator, we scale up our experiment to evaluate three LLMs using more data and generating five counterfactuals per explanation. For the automatic evaluation, 50 explanations along with 5 generated counterfactuals each (250 explanation, counterfactual pairs total) were evaluated.

\section{Results}

\subsection{Intermediate results}

\paragraph{GPT-4 Turbo is able to parse explanations well for summarization but not for medical suggestion.} Annotators find that the LLM successfully breaks down explanations with 96\% and 57\% accuracy for news summarization and medical suggestion respectively. For medical suggestion errors, the LLM either fails to identify a key detail or extracts an atomic unit incorrectly (extracted a unit it should not have or classified it as the incorrect type of information). Further investigation reveals that errors often involve minor details whereas the main points are successfully extracted (see Appendix \ref{app:explanation_parsing}). 

\paragraph{GPT-4 Turbo is able to generate simulatable counterfactuals for summarization but not for medical suggestion.} While almost all ($74/76$) generated counterfactuals are deemed simulatable in the summarization setting, only slightly more than half are for medical suggestion ($52/90$). Most of these non-simulatable counterfactuals contain $0.4-0.8$ of the atomic units of the explanation, indicating that counterfactual generation is more challenging in this setting. The distribution of generated counterfactuals, sorted by the proportion of atomic units that appear is shown in Appendix \ref{app:cf_gen_simulatability}.

\iffalse

\begin{table}[h]
\centering
\small
\renewcommand{\arraystretch}{1.3} 
\setlength{\tabcolsep}{12pt}
\begin{tabular}{|l|l|c|c|}
\hline
\textbf{Task} & \textbf{Explanation Method} & \textbf{Generality} & \textbf{Precision} \\
\hline
\multirow{2}{*}{News Summarization} & Chain of Thought & 0.52 & 0.81 \\
 & Post Hoc & 0.49 & 0.89 \\
\hline
\multirow{2}{*}{Medical Suggestion} & Chain of Thought & 0.20 & 0.51 \\
 & Post Hoc & 0.26 & 0.59 \\
\hline
\end{tabular}
\caption{Generality and precision across explanation types and tasks for human evaluation.}
\label{tab:human_eval_results}
\end{table}

\fi

\iffalse

\begin{table}[h]
\centering
\renewcommand{\arraystretch}{1.3}
\resizebox{\columnwidth}{!}{%
\begin{tabular}{|l|l|c|c|}
\hline
\textbf{Task} & \textbf{Explanation} & \textbf{Generality} & \textbf{Precision} \\
\hline
\multirow{2}{*}{\begin{tabular}[c]{@{}l@{}}News\\Summarization\end{tabular}} & CoT & 0.52 & 0.81 \\
\cline{2-4}
 & Post-hoc & 0.49 & 0.89 \\
\hline
\multirow{2}{*}{\begin{tabular}[c]{@{}l@{}}Medical\\Suggestion\end{tabular}} & CoT & 0.20 & 0.51 \\
\cline{2-4}
 & Post-hoc & 0.26 & 0.59 \\
\hline
\end{tabular}%
}
\caption{Generality and precision scores for GPT-4 Turbo from the human evaluation.}
\label{tab:human_eval_results}
\end{table}

\fi

\subsection{Human evaluation}

GPT-4 Turbo explanations lead to generalizable counterfactuals and consistent mental models in the case of summarization, with approximately $0.8$ of the inferred information appearing in the counterfactual output. In contrast, explanations in the medical suggestion setting are less generalizable and less precise (approximately $0.5$), indicating that LLMs may struggle more to reliably explain their behavior for this task. Chain-of-thought and Post-hoc explanations lead to similar results (see Table \ref{tab:human_eval_results}). For each pair of settings, we fit an independent sample t-test and find that the difference in both metrics when compared across tasks are significant ($p<0.05$), but not when compared across explanation types within the same task.

\begin{table}[h]
\centering
\setlength{\extrarowheight}{4pt} % Add vertical space above each row
\setlength{\arrayrulewidth}{0.8pt} % Increase border thickness (default is 0.4pt)
\resizebox{\columnwidth}{!}{%
\normalsize 
\begin{tabular}{|l|l|c|c|}
\hline
\multicolumn{1}{|c|}{\textbf{Task}} & \textbf{Explanation} & \textbf{Generality} & \textbf{Precision} \\[0.4em]
\hline
\multirow{4}{*}{\begin{tabular}[c]{@{}l@{}}News\\Summarization\end{tabular}} & \multirow{2}{*}{\begin{tabular}[c]{@{}l@{}}Chain-of-\\thought\end{tabular}} & \multirow{2}{*}{0.52} & \multirow{2}{*}{0.81} \\[0.3em]
& & & \\
\cline{2-4}
 & \multirow{2}{*}{Post-hoc} & \multirow{2}{*}{0.49} & \multirow{2}{*}{0.89} \\
 & & & \\
\hline
\multirow{4}{*}{\begin{tabular}[c]{@{}l@{}}Medical\\Suggestion\end{tabular}} & \multirow{2}{*}{\begin{tabular}[c]{@{}l@{}}Chain-of-\\thought\end{tabular}} & \multirow{2}{*}{0.20} & \multirow{2}{*}{0.51} \\[0.3em]
& & & \\
\cline{2-4}
 & \multirow{2}{*}{\large{Post-hoc}} & \multirow{2}{*}{0.26} & \multirow{2}{*}{0.59} \\
 & & & \\
\hline
\end{tabular}%
}
\caption{Generality and precision scores for GPT-4 Turbo from the human evaluation.}
\label{tab:human_eval_results}
\end{table}

As a sanity check for the pipeline, we generate counterfactual outputs conditioned on the original explanation and verify that these result in a precision score of $1.00$. This validates that our framework captures how well models follow their explanations. Therefore, the low precision scores for medical suggestion are due to the model's behavior not adhering to its explanation on that task rather than the evaluation setup.

\iffalse

\begin{table*}[ht]
\centering
\small
\renewcommand{\arraystretch}{1.5}
\resizebox{\textwidth}{!}{%
\begin{tabular}{l@{\hspace{2em}}c@{\hspace{2em}}cccc@{\hspace{1em}}cccc}
\toprule
\multirow{2}{*}{Task} & \multirow{2}{*}{\centering \quad Model} & \multicolumn{4}{c}{Chain-of-thought} & \multicolumn{4}{c}{Post-hoc} \\
\cmidrule(lr){3-6} \cmidrule(lr){7-10}
& & \# Expl & \# Samples & Generality & Precision & \# Expl & \# Samples & Generality & Precision \\
\midrule
\multirow{3}{*}[-1ex]{\makecell[l]{News\\Summarization}} 
& \makecell[c]{Claude 3.7 Sonnet} & 48 & 189 & 0.67 & 0.93 & 45 & 173 & 0.62 & 0.84 \\
\cmidrule(lr){2-10}
& \makecell[c]{GPT-4} & 46 & 160 & 0.59 & 0.84 & 48 & 193 & 0.65 & 0.78\\
\cmidrule(lr){2-10}
& \makecell[c]{Llama 3} & 47 & 172 & 0.67 & 0.74 & 43 & 153 & 0.67 & 0.66 \\
\midrule
\multirow{3}{*}[-1ex]{\makecell[l]{Medical\\Suggestion}} 
& \makecell[c]{Claude 3.7 Sonnet} & 24 & 55 & 0.21 & 0.48 & 24 & 78 & 0.20 & 0.66 \\
\cmidrule(lr){2-10}
& \makecell[c]{GPT-4} & 36 & 103 & 0.20 & 0.46 & 36 & 122 & 0.19 & 0.65 \\
\cmidrule(lr){2-10}
& \makecell[c]{Llama 3} & 24 & 69 & 0.19 & 0.56 & 30 & 110 & 0.20 & 0.66 \\
\bottomrule
\end{tabular}%
}
\caption{Generality and precision metrics across models, explanation types, and tasks from the automatic evaluation.}
\label{tab:automatic_eval}
\end{table*}

\fi

\begin{table*}[ht]
\centering
\normalsize
\setlength{\tabcolsep}{8pt}
\setlength{\arrayrulewidth}{0.8pt}
\renewcommand{\arraystretch}{1.8}
\resizebox{\textwidth}{!}{%
\begin{tabular}{|l|c|cccc|cccc|}
\hline
\multirow{2}{*}{\qquad \textbf{Task}} & \multirow{2}{*}{\centering \textbf{Model}} & \multicolumn{4}{c|}{\textbf{Chain-of-thought}} & \multicolumn{4}{c|}{\textbf{Post-hoc}} \\
\cline{3-10}
& & \# Expl & \# Samples & Generality & Precision & \# Expl & \# Samples & Generality & Precision \\
\hline
\multirow{3}{*}[-1ex]{\makecell[l]{News\\Summarization}} 
& \makecell[c]{Claude 3.7 Sonnet} & 48 & 189 & 0.67 & 0.93 & 45 & 173 & 0.62 & 0.84 \\
\cline{2-10}
& \makecell[c]{GPT-4} & 46 & 160 & 0.59 & 0.84 & 48 & 193 & 0.65 & 0.78\\
\cline{2-10}
& \makecell[c]{Llama 3} & 47 & 172 & 0.67 & 0.74 & 43 & 153 & 0.67 & 0.66 \\
\hline
\multirow{3}{*}[-1ex]{\makecell[l]{Medical\\Suggestion}} 
& \makecell[c]{Claude 3.7 Sonnet} & 24 & 55 & 0.21 & 0.48 & 24 & 78 & 0.20 & 0.66 \\ 
\cline{2-10}
& \makecell[c]{GPT-4} & 36 & 103 & 0.20 & 0.46 & 36 & 122 & 0.19 & 0.65 \\
\cline{2-10}
& \makecell[c]{Llama 3} & 24 & 69 & 0.19 & 0.56 & 30 & 110 & 0.20 & 0.66 \\
\hline
\end{tabular}%
}
\caption{Generality and precision metrics across models, explanation types, and tasks from the automatic evaluation.}
\label{tab:automatic_eval}
\end{table*}

\subsection{Automatic evaluation}

\paragraph{GPT-4 Turbo is able to approximate human annotation for our tasks.} Table \ref{tab:inter-annotator-agreement} reports the pairwise Cohen's Kappa between each pair of annotators and the LLM. We find that overall, GPT-4 Turbo achieves similar inter-annotator agreement to humans, with an average of 0.61 for human-LLM pairs compared to 0.54 for human-human pairs in news summarization and 0.65 for human-LLM pairs compared to 0.74 for human-human pairs in medical suggestion. We calculate a two-sided p-value for ratings between all annotation pairs and find that the observed agreement is significant.

The results of the automatic evaluation, presented in Table \ref{tab:automatic_eval}, are in line with our findings from the human evaluation. Namely, models are better at accurately explaining their behavior for summarization compared to medical suggestion while also remaining general such that these explanations apply to diverse counterfactuals. Chain-of-thought explanations lead to better mental models for summarization, which may reflect the skill-based nature of this task. On the other hand, Post-hoc explanations lead to more precise explanations for medical suggestion. Models may also differ in their ability to describe their behavior in a generative setting, for instance, Claude 3.7 Sonnet demonstrates noticeably better precision scores for news summarization compared to other models. 

\section{Discussion}

\paragraph{LLM explanations can be helpful for skill-based generation tasks but may struggle for knowledge-based generation tasks.} While model explanations do enable users to infer pieces of information that will appear in counterfactual outputs in both settings, the mental models produced are more accurate for summarization compared to medical suggestion. Furthermore, summarization explanations, which employ a high-level approach to their explanation, are also more generalizable. This may suggest that LLMs are better able to describe their behavior for skill-based tasks, where users can reliably infer the elements that will appear in the outputs, compared to knowledge-based tasks, where users are less capable of inferring specific points (e.g., suggestions). Additionally, utilizing Chain-of-thought prompting, which aligns with the skill-based nature of summarization, may lead to more precise explanations. It is important to consider that our evaluation only examined news summarization and medical suggestion specifically. As such, the results may vary for other skill-based and knowledge-based tasks. 

These findings may reflect the predictability of model behavior on different tasks types. For medical suggestion, minor variances in the question may lead to very different answers. For instance, if a user describes ``\textit{I experience chest pain sometimes when I exercise.}'' a model might respond ``\textit{consider reducing exercise intensity, try warming up thoroughly before exercise ...}''. However, if a user changes their question slightly to ``\textit{I experience chest pain sometimes when I exercise. I started taking a new pre-workout supplement with high caffeine recently.}'', although they are expressing the same symptoms, because of the presence of additional information the model might respond ``\textit{discontinue the supplement or use smaller dosages, the chest pain is likely related to caffeine-induced heart palpitations ...}''. These differences in the sensitivity of the model's output relative to the input may lead to the observed differences in precision and generality in our experiments.

\paragraph{Our counterfactual simulatability evaluation framework is effective in the summarization setting but less suited for medical suggestion.} Although LLMs are able to automate the human annotation steps in our evaluation, they demonstrate issues in other aspects for the medical suggestion task. Specifically, our annotators identify many errors in the parsing of explanations into atomic units, where the LLM misses key information or mis-classifies the information. In one instance, ``possible heat-induced asthma'' is incorrectly extracted as a key symptom when in reality it is a potential cause. Unlike the summarization setting, this added requirement of classifying the type of information introduces more complexity, making the LLM a less effective tool in the evaluation pipeline. Additionally, we found that the LLM is unable to produce as many simulatable counterfactuals in the medical domain compared to summarization. There is significant room for improvement towards adapting counterfactual simulatability for knowledge-based tasks like medical suggestion, where the explanation and decision process for an LLM is less explicit and relies heavily on knowledge encoded in the model.

\section{Limitations}

First, the generality metric relies on our LLM (GPT-4 Turbo) being able to generate diverse, yet simulatable counterfactuals. Although we briefly experimented with counterfactual generation, more work can be done to assess whether other models or prompting strategies may lead to higher quality counterfactuals. Additionally, cosine similarity might not be the optimal way to judge diversity across different tasks as it is heavily content-based. For example, while many counterfactuals for summarization followed the same structure (e.g., a headline followed by when/where it occurred followed by a quote ...), they were still rewarded with high generality because the specific topics and words differed. 

Second, in a few cases there was ambiguity when determining whether atomic units were \lq present\rq in the counterfactual/the counterfactual output. Since these units of information may appear with different wording or even be implied, we encouraged annotators to use their best judgement and leave notes. We leave investigation into ways of improving this matching to future work. 

Finally, we focused on demonstrating our framework using explanations generated from a few popular LLMs. Testing newer types of models (e.g., reasoning models) and different task setups (e.g., multiple interactions) are a promising direction. Furthermore, expanding the set of metrics to consider aspects such as explanation comprehensiveness, faithfulness, and plausibility, and assessing how they vary with counterfactual simulatability, may also be worthwhile. 

\paragraph*{Supplementary Materials Availability Statement:} Our source code as well as an anonymized version of the data used in the human evaluation is made available at \url{github.com/mlimpijankit/counterfactual-simulatability-generation-tasks}.

\section{Acknowledgments}

This research is supported in part by the Office of the Director of National Intelligence (ODNI), Intelligence Advanced Research Projects Activity (IARPA), via the HIATUS Program contract \#2022-22072200005 and the National Science Foundation and by DoD OUSD (R\&E) under Cooperative Agreement PHY-2229929 (The NSF AI Institute for Artificial and Natural Intelligence). The views, opinions and/or findings expressed are those of the author and should not be interpreted as representing the official views or policies of the Department of Defense, the National Science Foundation or the U.S. Government. One of the authors holds an equity interest in OpenAI.

\bibliography{custom}

@inproceedings{chen-simulatbility,
author = {Chen, Yanda and Zhong, Ruiqi and Ri, Narutatsu and Zhao, Chen and He, He and Steinhardt, Jacob and Yu, Zhou and McKeown, Kathleen},
title = {Do models explain themselves? counterfactual simulatability of natural language explanations},
year = {2024},
publisher = {JMLR.org},
booktitle = {Proceedings of the 41st International Conference on Machine Learning},
articleno = {310},
numpages = {25},
location = {Vienna, Austria},
series = {ICML'24}
}

@article{ji2023survey,
  title={Survey of hallucination in natural language generation},
  author={Ji, Ziwei and Lee, Nayeon and Frieske, Rita and Yu, Tiezheng and Su, Dan and Xu, Yan and Ishii, Etsuko and Bang, Ye Jin and Madotto, Andrea and Fung, Pascale},
  journal={ACM Computing Surveys},
  volume={55},
  number={12},
  pages={1--38},
  year={2023},
  publisher={ACM New York, NY}
}

@inproceedings{chen-etal-2022-learning-generate,
    title = "Learning to Generate Explanation from e-Hospital Services for Medical Suggestion",
    author = "Chen, Wei-Lin  and
      Yen, An-Zi  and
      Huang, Hen-Hsen  and
      Chen, Hsin-Hsi",
    editor = "Calzolari, Nicoletta  and
      Huang, Chu-Ren  and
      Kim, Hansaem  and
      Pustejovsky, James  and
      Wanner, Leo  and
      Choi, Key-Sun  and
      Ryu, Pum-Mo  and
      Chen, Hsin-Hsi  and
      Donatelli, Lucia  and
      Ji, Heng  and
      Kurohashi, Sadao  and
      Paggio, Patrizia  and
      Xue, Nianwen  and
      Kim, Seokhwan  and
      Hahm, Younggyun  and
      He, Zhong  and
      Lee, Tony Kyungil  and
      Santus, Enrico  and
      Bond, Francis  and
      Na, Seung-Hoon",
    booktitle = "Proceedings of the 29th International Conference on Computational Linguistics",
    month = oct,
    year = "2022",
    address = "Gyeongju, Republic of Korea",
    publisher = "International Committee on Computational Linguistics",
    url = "https://aclanthology.org/2022.coling-1.260/",
    pages = "2946--2951"
}

@inproceedings{kunz-kuhlmann-2024-properties,
    title = "Properties and Challenges of {LLM}-Generated Explanations",
    author = "Kunz, Jenny  and
      Kuhlmann, Marco",
    editor = "Blodgett, Su Lin  and
      Cercas Curry, Amanda  and
      Dev, Sunipa  and
      Madaio, Michael  and
      Nenkova, Ani  and
      Yang, Diyi  and
      Xiao, Ziang",
    booktitle = "Proceedings of the Third Workshop on Bridging Human--Computer Interaction and Natural Language Processing",
    month = jun,
    year = "2024",
    address = "Mexico City, Mexico",
    publisher = "Association for Computational Linguistics",
    url = "https://aclanthology.org/2024.hcinlp-1.2/",
    doi = "10.18653/v1/2024.hcinlp-1.2",
    pages = "13--27"
}

@inproceedings{sheng-etal-2021-societal,
    title = "Societal Biases in Language Generation: Progress and Challenges",
    author = "Sheng, Emily  and
      Chang, Kai-Wei  and
      Natarajan, Prem  and
      Peng, Nanyun",
    editor = "Zong, Chengqing  and
      Xia, Fei  and
      Li, Wenjie  and
      Navigli, Roberto",
    booktitle = "Proceedings of the 59th Annual Meeting of the Association for Computational Linguistics and the 11th International Joint Conference on Natural Language Processing (Volume 1: Long Papers)",
    month = aug,
    year = "2021",
    address = "Online",
    publisher = "Association for Computational Linguistics",
    url = "https://aclanthology.org/2021.acl-long.330/",
    doi = "10.18653/v1/2021.acl-long.330",
    pages = "4275--4293"
}

@inproceedings{madsen-etal-2024-self,
    title = "Are self-explanations from Large Language Models faithful?",
    author = "Madsen, Andreas  and
      Chandar, Sarath  and
      Reddy, Siva",
    editor = "Ku, Lun-Wei  and
      Martins, Andre  and
      Srikumar, Vivek",
    booktitle = "Findings of the Association for Computational Linguistics: ACL 2024",
    month = aug,
    year = "2024",
    address = "Bangkok, Thailand",
    publisher = "Association for Computational Linguistics",
    url = "https://aclanthology.org/2024.findings-acl.19/",
    doi = "10.18653/v1/2024.findings-acl.19",
    pages = "295--337"
}

@book{gentner-mental,
  author = {Gentner, Dedre and Stevens, Albert L.},
  title = {Mental Models},
  ISBN = {9781317769408},
  url = {http://dx.doi.org/10.4324/9781315802725},
  DOI = {10.4324/9781315802725},
  publisher = {Psychology Press},
  year = {2014},
  month = jan 
}

@article{payne-mental,
    author = {Stephen J. Payne},
    title = {A descriptive study of mental models†},
    journal = {Behaviour \& Information Technology},
    volume = {10},
    number = {1},
    pages = {3--21},
    year = {1991},
    publisher = {Taylor \& Francis},
    doi = {10.1080/01449299108924268},
    URL = {https://doi.org/10.1080/01449299108924268},
    eprint = {https://doi.org/10.1080/01449299108924268}
}

@inbook{du-voice,
  title = {Voice User Interface Interaction Design Research Based on User Mental Model in Autonomous Vehicle},
  ISBN = {9783319912509},
  ISSN = {1611-3349},
  url = {http://dx.doi.org/10.1007/978-3-319-91250-9_10},
  DOI = {10.1007/978-3-319-91250-9_10},
  booktitle = {Human-Computer Interaction. Interaction Technologies},
  publisher = {Springer International Publishing},
  author = {Du,  Yuemeng and Qin,  Jingyan and Zhang,  Shujing and Cao,  Sha and Dou,  Jinhua},
  year = {2018},
  pages = {117–132}
}

@inproceedings{vasconcelos-cost,
  title={When do XAI methods work? A cost-benefit approach to human-AI collaboration},
  author={Vasconcelos, Helena and J{\"o}rke, Matthew and Grunde-McLaughlin, Madeleine and Krishna, Ranjay and Gerstenberg, Tobias and Bernstein, Michael S},
  booktitle={CHI Workshop on Trust and Reliance in AI-Human Teams},
  pages={1--15},
  year={2022}
}

@article{senoner-collab,
  title = {Explainable AI improves task performance in human–AI collaboration},
  volume = {14},
  ISSN = {2045-2322},
  url = {http://dx.doi.org/10.1038/s41598-024-82501-9},
  DOI = {10.1038/s41598-024-82501-9},
  number = {1},
  journal = {Scientific Reports},
  publisher = {Springer Science and Business Media LLC},
  author = {Senoner,  Julian and Schallmoser,  Simon and Kratzwald,  Bernhard and Feuerriegel,  Stefan and Netland,  Torbjørn},
  year = {2024},
  month = dec 
}

@article{merry-xai,
  title = {A mental models approach for defining explainable artificial intelligence},
  volume = {21},
  ISSN = {1472-6947},
  url = {http://dx.doi.org/10.1186/s12911-021-01703-7},
  DOI = {10.1186/s12911-021-01703-7},
  number = {1},
  journal = {BMC Medical Informatics and Decision Making},
  publisher = {Springer Science and Business Media LLC},
  author = {Merry,  Michael and Riddle,  Pat and Warren,  Jim},
  year = {2021},
  month = dec 
}

@inbook{lei-mobile,
  title = {The Influence of Matching Degree of the User’s Inherent Mental Model and the Product’s Embedded Mental Model on the Mobile User Experience},
  ISBN = {9783319395166},
  ISSN = {1611-3349},
  url = {http://dx.doi.org/10.1007/978-3-319-39516-6_31},
  DOI = {10.1007/978-3-319-39516-6_31},
  booktitle = {Human-Computer Interaction. Interaction Platforms and Techniques},
  publisher = {Springer International Publishing},
  author = {Lei,  Tian and Liu,  Xu and Wu,  Lei and Jin,  Ziliang and Wang,  Yuhui and Wei,  Shuaili},
  year = {2016},
  pages = {320–329}
}

@article{bansal-accuracy,
  title = {Beyond Accuracy: The Role of Mental Models in Human-AI Team Performance},
  volume = {7},
  ISSN = {2769-1330},
  url = {http://dx.doi.org/10.1609/hcomp.v7i1.5285},
  DOI = {10.1609/hcomp.v7i1.5285},
  journal = {Proceedings of the AAAI Conference on Human Computation and Crowdsourcing},
  publisher = {Association for the Advancement of Artificial Intelligence (AAAI)},
  author = {Bansal,  Gagan and Nushi,  Besmira and Kamar,  Ece and Lasecki,  Walter S. and Weld,  Daniel S. and Horvitz,  Eric},
  year = {2019},
  month = oct,
  pages = {2–11}
}

@inproceedings{rutjes-xai,
    title = "Considerations on explainable AI and users{\textquoteright} mental models",
    author = "Heleen Rutjes and Martijn Willemsen and Wijnand IJsselsteijn",
    year = "2019",
    month = may,
    day = "4",
    language = "English",
    booktitle = "Where is the Human? Bridging the Gap Between AI and HCI",
    publisher = "Association for Computing Machinery, Inc",
    address = "United States",
    note = "CHI 2019 Workshop : Where is the Human? Bridging the Gap Between AI and HCI ; Conference date: 04-05-2019 Through 04-05-2019",
}

@article{steyvers-think,
  title = {What large language models know and what people think they know},
  ISSN = {2522-5839},
  url = {http://dx.doi.org/10.1038/s42256-024-00976-7},
  DOI = {10.1038/s42256-024-00976-7},
  journal = {Nature Machine Intelligence},
  publisher = {Springer Science and Business Media LLC},
  author = {Steyvers,  Mark and Tejeda,  Heliodoro and Kumar,  Aakriti and Belem,  Catarina and Karny,  Sheer and Hu,  Xinyue and Mayer,  Lukas W. and Smyth,  Padhraic},
  year = {2025},
  month = jan 
}

@inproceedings{nallapati-etal-2016-abstractive,
    title = "Abstractive Text Summarization using Sequence-to-sequence {RNN}s and Beyond",
    author = "Nallapati, Ramesh  and
      Zhou, Bowen  and
      dos Santos, Cicero  and
      Gu{\ensuremath{\dot{}}}l{\c{c}}ehre, {\c{C}}a{\u{g}}lar  and
      Xiang, Bing",
    editor = "Riezler, Stefan  and
      Goldberg, Yoav",
    booktitle = "Proceedings of the 20th {SIGNLL} Conference on Computational Natural Language Learning",
    month = aug,
    year = "2016",
    address = "Berlin, Germany",
    publisher = "Association for Computational Linguistics",
    url = "https://aclanthology.org/K16-1028/",
    doi = "10.18653/v1/K16-1028",
    pages = "280--290"
}

@InProceedings{michalowski-explain,
author="Michalowski, Martin
and Wilk, Szymon
and Bauer, Jenny M.
and Carrier, Marc
and Delluc, Aurelien
and Le Gal, Gr{\'e}goire
and Wang, Tzu-Fei
and Siegal, Deborah
and Michalowski, Wojtek",
editor="Finkelstein, Joseph
and Moskovitch, Robert
and Parimbelli, Enea",
title="Manually-Curated Versus LLM-Generated Explanations for Complex Patient Cases: An Exploratory Study with Physicians",
booktitle="Artificial Intelligence in Medicine",
year="2024",
publisher="Springer Nature Switzerland",
address="Cham",
pages="313--323",
isbn="978-3-031-66535-6"
}

@inproceedings{wei-cot,
author = {Wei, Jason and Wang, Xuezhi and Schuurmans, Dale and Bosma, Maarten and Ichter, Brian and Xia, Fei and Chi, Ed H. and Le, Quoc V. and Zhou, Denny},
title = {Chain-of-thought prompting elicits reasoning in large language models},
year = {2022},
isbn = {9781713871088},
publisher = {Curran Associates Inc.},
address = {Red Hook, NY, USA},
booktitle = {Proceedings of the 36th International Conference on Neural Information Processing Systems},
articleno = {1800},
numpages = {14},
location = {New Orleans, LA, USA},
series = {NIPS '22}
}

@inproceedings{camburu-esnli,
 author = {Camburu, Oana-Maria and Rockt\"{a}schel, Tim and Lukasiewicz, Thomas and Blunsom, Phil},
 booktitle = {Advances in Neural Information Processing Systems},
 editor = {S. Bengio and H. Wallach and H. Larochelle and K. Grauman and N. Cesa-Bianchi and R. Garnett},
 pages = {},
 publisher = {Curran Associates, Inc.},
 title = {e-SNLI: Natural Language Inference with Natural Language Explanations},
 url = {https://proceedings.neurips.cc/paper_files/paper/2018/file/4c7a167bb329bd92580a99ce422d6fa6-Paper.pdf},
 volume = {31},
 year = {2018}
}

@misc{huang-explain,
      title={Can Large Language Models Explain Themselves? A Study of LLM-Generated Self-Explanations}, 
      author={Shiyuan Huang and Siddarth Mamidanna and Shreedhar Jangam and Yilun Zhou and Leilani H. Gilpin},
      year={2023},
      eprint={2310.11207},
      archivePrefix={arXiv},
      primaryClass={cs.CL},
      url={https://arxiv.org/abs/2310.11207}, 
}

@inproceedings{deyoung-eraser,
    title = "{ERASER}: {A} Benchmark to Evaluate Rationalized {NLP} Models",
    author = "DeYoung, Jay  and
      Jain, Sarthak  and
      Rajani, Nazneen Fatema  and
      Lehman, Eric  and
      Xiong, Caiming  and
      Socher, Richard  and
      Wallace, Byron C.",
    editor = "Jurafsky, Dan  and
      Chai, Joyce  and
      Schluter, Natalie  and
      Tetreault, Joel",
    booktitle = "Proceedings of the 58th Annual Meeting of the Association for Computational Linguistics",
    month = jul,
    year = "2020",
    address = "Online",
    publisher = "Association for Computational Linguistics",
    url = "https://aclanthology.org/2020.acl-main.408/",
    doi = "10.18653/v1/2020.acl-main.408",
    pages = "4443--4458"
}

@misc{fayyaz-alignment,
      title={Evaluating Human Alignment and Model Faithfulness of LLM Rationale}, 
      author={Mohsen Fayyaz and Fan Yin and Jiao Sun and Nanyun Peng},
      year={2024},
      eprint={2407.00219},
      archivePrefix={arXiv},
      primaryClass={cs.CL},
      url={https://arxiv.org/abs/2407.00219}, 
}

@article{zhichao-scrutability,
author = {Xu, Zhichao and Zeng, Hansi and Tan, Juntao and Fu, Zuohui and Zhang, Yongfeng and Ai, Qingyao},
title = {A Reusable Model-agnostic Framework for Faithfully Explainable Recommendation and System Scrutability},
year = {2023},
issue_date = {January 2024},
publisher = {Association for Computing Machinery},
address = {New York, NY, USA},
volume = {42},
number = {1},
issn = {1046-8188},
url = {https://doi.org/10.1145/3605357},
doi = {10.1145/3605357},
journal = {ACM Trans. Inf. Syst.},
month = aug,
articleno = {29},
numpages = {29}
}

@inproceedings{krishna-posthoc,
 author = {Krishna, Satyapriya and Ma, Jiaqi and Slack, Dylan and Ghandeharioun, Asma and  Singh, Sameer and Lakkaraju, Himabindu},
 booktitle = {Advances in Neural Information Processing Systems},
 editor = {A. Oh and T. Naumann and A. Globerson and K. Saenko and M. Hardt and S. Levine},
 pages = {65468--65483},
 publisher = {Curran Associates, Inc.},
 title = {Post Hoc Explanations of Language Models Can Improve Language Models},
 url = {https://proceedings.neurips.cc/paper_files/paper/2023/file/ce65173b994cf7c925c71b482ee14a8d-Paper-Conference.pdf},
 volume = {36},
 year = {2023}
}

@inproceedings{turpin-cot,
author = {Turpin, Miles and Michael, Julian and Perez, Ethan and Bowman, Samuel R.},
title = {Language models don't always say what they think: unfaithful explanations in chain-of-thought prompting},
year = {2023},
publisher = {Curran Associates Inc.},
address = {Red Hook, NY, USA},
booktitle = {Proceedings of the 37th International Conference on Neural Information Processing Systems},
articleno = {3275},
numpages = {14},
location = {New Orleans, LA, USA},
series = {NIPS '23}
}

@inproceedings{zhou-commonsense,
    title = "Probing Commonsense Explanation in Dialogue Response Generation",
    author = "Zhou, Pei  and
      Jandaghi, Pegah  and
      Cho, Hyundong  and
      Lin, Bill Yuchen  and
      Pujara, Jay  and
      Ren, Xiang",
    editor = "Moens, Marie-Francine  and
      Huang, Xuanjing  and
      Specia, Lucia  and
      Yih, Scott Wen-tau",
    booktitle = "Findings of the Association for Computational Linguistics: EMNLP 2021",
    month = nov,
    year = "2021",
    address = "Punta Cana, Dominican Republic",
    publisher = "Association for Computational Linguistics",
    url = "https://aclanthology.org/2021.findings-emnlp.349/",
    doi = "10.18653/v1/2021.findings-emnlp.349",
    pages = "4132--4146"
}

@inproceedings{
ho-wikiwhy,
title={WikiWhy: Answering and Explaining Cause-and-Effect Questions},
author={Matthew Ho and Aditya Sharma and Justin Chang and Michael Saxon and Sharon Levy and Yujie Lu and William Yang Wang},
booktitle={The Eleventh International Conference on Learning Representations },
year={2023},
url={https://openreview.net/forum?id=vaxnu-Utr4l}
}

@inproceedings{gao-dialogue,
    title = "Self-Explanation Prompting Improves Dialogue Understanding in Large Language Models",
    author = "Gao, Haoyu  and
      Lin, Ting-En  and
      Li, Hangyu  and
      Yang, Min  and
      Wu, Yuchuan  and
      Ma, Wentao  and
      Huang, Fei  and
      Li, Yongbin",
    editor = "Calzolari, Nicoletta  and
      Kan, Min-Yen  and
      Hoste, Veronique  and
      Lenci, Alessandro  and
      Sakti, Sakriani  and
      Xue, Nianwen",
    booktitle = "Proceedings of the 2024 Joint International Conference on Computational Linguistics, Language Resources and Evaluation (LREC-COLING 2024)",
    month = may,
    year = "2024",
    address = "Torino, Italia",
    publisher = "ELRA and ICCL",
    url = "https://aclanthology.org/2024.lrec-main.1269/",
    pages = "14567--14578"
}

@inproceedings{fragkathoulas-qa,
author = {Fragkathoulas, Christos and Chlapanis, Odysseas Spyridon},
title = {Local Explanations and Self-Explanations for Assessing Faithfulness in black-box LLMs},
year = {2024},
isbn = {9798400709821},
publisher = {Association for Computing Machinery},
address = {New York, NY, USA},
url = {https://doi.org/10.1145/3688671.3688775},
doi = {10.1145/3688671.3688775},
booktitle = {Proceedings of the 13th Hellenic Conference on Artificial Intelligence},
articleno = {56},
numpages = {5},
location = {
},
series = {SETN '24}
}

@inproceedings{lyu-faithful,
    title = "Faithful Chain-of-Thought Reasoning",
    author = "Lyu, Qing  and
      Havaldar, Shreya  and
      Stein, Adam  and
      Zhang, Li  and
      Rao, Delip  and
      Wong, Eric  and
      Apidianaki, Marianna  and
      Callison-Burch, Chris",
    editor = "Park, Jong C.  and
      Arase, Yuki  and
      Hu, Baotian  and
      Lu, Wei  and
      Wijaya, Derry  and
      Purwarianti, Ayu  and
      Krisnadhi, Adila Alfa",
    booktitle = "Proceedings of the 13th International Joint Conference on Natural Language Processing and the 3rd Conference of the Asia-Pacific Chapter of the Association for Computational Linguistics (Volume 1: Long Papers)",
    month = nov,
    year = "2023",
    address = "Nusa Dua, Bali",
    publisher = "Association for Computational Linguistics",
    url = "https://aclanthology.org/2023.ijcnlp-main.20/",
    doi = "10.18653/v1/2023.ijcnlp-main.20",
    pages = "305--329"
}

@article{wright2022generating,
  title={Generating scientific claims for zero-shot scientific fact checking},
  author={Wright, Dustin and Wadden, David and Lo, Kyle and Kuehl, Bailey and Cohan, Arman and Augenstein, Isabelle and Wang, Lucy Lu},
  journal={arXiv preprint arXiv:2203.12990},
  year={2022}
}

@article{chen2022generating,
  title={Generating literal and implied subquestions to fact-check complex claims},
  author={Chen, Jifan and Sriram, Aniruddh and Choi, Eunsol and Durrett, Greg},
  journal={arXiv preprint arXiv:2205.06938},
  year={2022}
}

@article{kamoi2023wice,
  title={Wice: Real-world entailment for claims in wikipedia},
  author={Kamoi, Ryo and Goyal, Tanya and Rodriguez, Juan Diego and Durrett, Greg},
  journal={arXiv preprint arXiv:2303.01432},
  year={2023}
}

@article{wojciechowski2024faithful,
  title={Faithful and plausible natural language explanations for image classification: A pipeline approach},
  author={Wojciechowski, Adam and Lango, Mateusz and Dusek, Ondrej},
  journal={arXiv preprint arXiv:2407.20899},
  year={2024}
}

\clearpage

\appendix

\section{Explanation Decomposition}
\label{app:explanation_parsing}

We categorize the errors reported by our annotators for explanation decomposition using GPT-4 Turbo during the human evaluation. Examples of each type of error are also provided below. Note that errors are subjective and often involve small details.

\begin{table}[h!]
\centering
\resizebox{\columnwidth}{!}{%
\begin{tabular}{lcc}
\hline
\textbf{Parsed Explanations} & \textbf{News Summarization} & \textbf{Medical Suggestion} \\
 & \textbf{($n=26$)} & \textbf{($n=30$)} \\
\hline
Accuracy & 0.96 & 0.57 \\
\hline
\multicolumn{3}{l}{\textit{Breakdown of Incorrect Examples:}} \\
\quad Missing Extraction & 1 & 8 \\
\quad Incorrect Extraction & 0 & 4 \\
\quad Missing and Incorrect Extraction & 0 & 1 \\
\hline
\end{tabular}%
}
\caption{Explanation parsing errors across tasks.}
\label{tab:explanation_parsing_errors}
\end{table}

\subsection*{Example of a Missing Extraction}

\textit{In the example below, ``anxiety and stress'' are mentioned in the explanation but not extracted.}

\paragraph{Explanation.} The symptoms you describe, including chest tightness, difficulty breathing, and the presence of white phlegm, along with discomfort in the stomach and worsening symptoms throughout the day, suggest a complex interaction between gastrointestinal and possibly psychosomatic factors. Superficial gastritis diagnosed previously might not fully explain the severity and range of your symptoms, especially if the treatment provided little relief. The persistence and nature of your symptoms indicate the need for a more comprehensive gastrointestinal evaluation to rule out other conditions such as gastroesophageal reflux disease (GERD), which can cause acid reflux, chest discomfort, and respiratory symptoms. Additionally, the significant impact of anxiety and stress on your physical health suggests that addressing these psychological factors is crucial. Cognitive behavioral therapy (CBT) is effective in managing anxiety and may reduce the intensity of the symptoms by helping you develop better coping mechanisms.

\paragraph{GPT-4 Turbo Parsed Explanation (Symptoms).}
\begin{itemize}
    \item chest tightness and difficulty breathing
    \item presence of white phlegm
    \item discomfort in the stomach
    \item symptoms worsen throughout the day
    \item previously diagnosed with superficial gastritis
    \item treatment for gastritis provided little relief
    \item \texttt{Missing Extraction: anxiety and stress}
\end{itemize}

\paragraph{GPT-4 Turbo Parsed Explanation (Suggestions).}
\begin{itemize}
    \item conduct a comprehensive gastrointestinal evaluation to rule out conditions like GERD 
    \item address psychological factors such as anxiety and stress 
    \item consider Cognitive Behavioral Therapy (CBT) to manage anxiety and develop coping mechanisms
\end{itemize}

\subsection*{Example of an Incorrect Extraction}

\textit{In the example below, ``possible heat-induced asthma or heat intolerance'' is incorrectly extracted in the symptoms category. The annotator noted that this was more of a potential cause rather than an expressed symptom.}

\paragraph{Explanation.} The patient experiences difficulty breathing, particularly during the summer or in hot temperatures, which improves significantly in cooler environments such as air-conditioned rooms or when using a fan. This pattern suggests that the symptoms may be related to heat-induced asthma or a condition known as heat intolerance. Heat can lead to increased airway resistance and trigger respiratory symptoms in sensitive individuals. It is also possible that environmental allergens, which are often more prevalent in warm weather, could be exacerbating the condition. To manage these symptoms, staying cool and avoiding heat exposure is crucial. However, a thorough evaluation by a healthcare professional is necessary to rule out other respiratory conditions and to confirm the diagnosis.

\paragraph{GPT-4 Turbo Parsed Explanation (Symptoms).}
\begin{itemize}
    \item difficulty breathing, especially during summer or in hot temperatures
    \item improvement in symptoms in cooler environments
    \item possible heat-induced asthma or heat intolerance (\texttt{Incorrect Extraction})
\end{itemize}

\paragraph{GPT-4 Turbo Parsed Explanation (Suggestions).}
\begin{itemize}
    \item stay cool and avoid heat exposure
    \item undergo a thorough evaluation by a healthcare professional to rule out other respiratory conditions and confirm the diagnosis
\end{itemize}

\section{Counterfactual Generation Simulatability}
\label{app:cf_gen_simulatability}

For each counterfactual in the human evaluation, we measure the extent to which they are simulatable by the proportion of atomic explanation units they contain. This distribution is displayed in Table \ref{tab:cf_gen_simulatability}.

\begin{table}[h!]
\centering
\resizebox{\columnwidth}{!}{%
\renewcommand{\arraystretch}{1.3}
\begin{tabular}{|l|r|r|}
\hline
\textbf{Proportion of atomic units present} & \multicolumn{1}{c|}{\textbf{News}} & \multicolumn{1}{c|}{\textbf{Medical}} \\
\textbf{in counterfactual (human annotated)} & \multicolumn{1}{c|}{\textbf{Summarization}} & \multicolumn{1}{c|}{\textbf{Suggestion}} \\
\hline
1.00* & 74 & 52 \\
\hline
0.80--0.99 & 2 & 0 \\
\hline
0.60--0.79 & 2 & 20 \\
\hline
0.40--0.59 & 0 & 17 \\
\hline
0.20--0.39 & 0 & 1 \\
\hline
0.00--0.19 & 0 & 0 \\
\hline
\hline
\textbf{Total} & \textbf{76} & \textbf{90} \\
\hline
\end{tabular}%
}
\caption{Distribution of the proportion of atomic units present in GPT4-Turbo generated counterfactuals across tasks. *Indicates the set of simulatable counterfactuals.}
\label{tab:cf_gen_simulatability}
\end{table}

\section{Counterfactual Generation Generality}
\label{app:cf_gen_generality}

We assess how generality and simulatability changes as the number of counterfactuals generated per explanation is increased. Since news summarizations are expensive to generate due to the prompt length, we only use 10 explanations in this experiment compared to 30 for medical suggestion. Note that in the summarization and 10 counterfactuals setting, not all 100 counterfactuals were generated due to errors parsing the LLMs output.

\begin{table}[H]
\centering
\resizebox{\columnwidth}{!}{%
\renewcommand{\arraystretch}{1.5}
\setlength{\tabcolsep}{10pt}
\begin{tabular}{p{5cm}>{\centering\arraybackslash}p{2cm}>{\centering\arraybackslash}p{2cm}>{\centering\arraybackslash}p{2cm}}
\hline
\multirow{2}{*}{\textbf{Metric}} & \multicolumn{3}{c}{\textbf{Counterfactuals generated per explanation}} \\
\cline{2-4}
 & \textbf{3} & \textbf{5} & \textbf{10} \\
\hline
\multicolumn{4}{l}{\textit{News Summarization}} \\
\hline
Generality & 0.497 & 0.512 & 0.595 \\
Simulatable counterfactuals & 18 & 30 & 38 \\
Total generated counterfactuals & 30 & 50 & 80 \\
\hline
\multicolumn{4}{l}{\textit{Medical Suggestion}} \\
\hline
Generality score & 0.187 & 0.218 & 0.227 \\
Simulatable counterfactuals & 41 & 66 & 95 \\
Total generated counterfactuals & 90 & 150 & 300 \\
\hline
\end{tabular}%
}
\caption{Generality and simulatability metrics as the number of generated counterfactuals per explanation increases.}
\end{table}

\section{LLM Prompts}
\label{app:llm_prompts}

We provide the prompts used in our evaluation pipeline.

\subsection*{News Summarization: Chain-of-thought explanations}

\begin{quote}
    Your task is to summarize the document provided. First, describe on a high level the essential elements that the summary should include. Crucially, your explanation should only mention high-level, abstract, and generalized themes, and MUST NOT leak the topic, actions, subjects, or any details of the document. You will be HEAVILY penalized if the explanation is too specific. See below examples for explanations. Following this, compose the summary. Your response should be in the format "Explanation:" followed by "Summary:". Closely follow the format in the examples below.

Example:

Document: (CNN)Share, and your gift will be multiplied. That may sound like an esoteric adage, but when Zully Broussard selflessly decided to give one of her kidneys to a stranger, her generosity paired up with big data. It resulted in six patients receiving transplants. That …

Explanation: Start by highlighting the subject's key decision or action. Next, identify the factors that contributed to the event's success to understand the underlying reasons. Then, outline the major outcomes and include aggregated statistics to provide a data-supported overview of the impact.

Summary: Zully Broussard decided to give a kidney to a stranger. A new computer program helped her donation spur transplants for six kidney patients.

Example:

Document: (CNN)On the 6th of April 1996, San Jose Clash and DC United strode out in front of 31,683 expectant fans at the Spartan Stadium in San Jose, California. The historic occasion was the first ever Major League Soccer match -- a brave new dawn for the …

Explanation: Start by identifying the milestone event and its specific date, as this is the focal point of the document. Next, trace the developments and changes that led up to this milestone, noting any significant dates. Additionally, include any ongoing debates or challenges related to the topic, which will help contextualize the current situation and offer insights into possible future developments.

Summary: The 20th MLS season begins this weekend. League has changed dramatically since its inception in 1996. Some question whether rules regarding salary caps and transfers need to change.

Example: ...

Example: ...

Your turn:

Document: \texttt{[DOCUMENT]}

\end{quote}

\subsection*{News Summarization: Post-hoc explanations}

\begin{quote}
Your task is to summarize the document provided. First, compose the summary. Following this, describe on a high level the essential elements that the summary should include. Crucially, your explanation should only mention high-level, abstract, and generalized themes, and MUST NOT leak the topic, actions, subjects, or any details of the document. You will be HEAVILY penalized if the explanation is too specific. See below examples for explanations. Following this, compose the summary. Your response should be in the format "Summary:" followed by "Explanation:". Closely follow the format in the examples below.

Example:

Document: (CNN)Share, and your gift will be multiplied. That may sound like an esoteric adage, but when Zully Broussard selflessly decided to give one of her kidneys to a stranger, her generosity paired up with big data. It resulted in six patients receiving transplants. That surprised and wowed her. "I …

Summary: Zully Broussard decided to give a kidney to a stranger. A new computer program helped her donation spur transplants for six kidney patients.

Explanation: Start by highlighting the subject's key decision or action. Next, identify the factors that contributed to the event's success to understand the underlying reasons. Then, outline the major outcomes and include aggregated statistics to provide a data-supported overview of the impact.

Example:

Document: (CNN)French striker Bafetimbi Gomis, who has a history of fainting, said he is now "feeling well" after collapsing during Swansea's 3-2 loss at Tottenham in the Premier League on Wednesday. The worrying incident occurred in the first half at White Hart Lane -- after Tottenham scored in the …

Summary: Bafetimbi Gomis collapses within 10 minutes of kickoff at Tottenham. But he reportedly left the pitch conscious and wearing an oxygen mask. Gomis later said that he was "feeling well" The incident came three years after Fabrice Muamba collapsed at White Hart Lane.

Explanation: Focus on the key event described in the document and specify when it occurred. Highlight important details, particularly those that are surprising or contradictory. Include relevant quotes or statements from individuals involved to add depth and authenticity. Also, reference related or prior events to give a fuller context, helping to situate the incident within a broader historical framework.

Example: ...

Your turn:

Document: \texttt{[DOCUMENT]}

\end{quote}

\subsection*{News Summarization: Counterfactual generation}

\begin{quote}
You will be asked to first read an AI's Decision Process for summarization. Then, you will be asked to craft 5 CNN-style news articles that you can confidently guess the AI's summary to based on its provided decision process. The content in the crafted news articles should be diverse. Do not use new lines in the crafted article. Start your crafted medical question with "Crafted Article 1:", "Crafted Article 2:" and so on. Closely follow the format in the examples below.

Example:

AI's Decision Process: Start by highlighting the subject's key decision or action. Next, identify the factors that contributed to the event's success to understand the underlying reasons. Then, outline the major outcomes and include aggregated statistics to provide a data-supported overview of the impact.

Crafted Article 1: (CNN)The world's battle against polio took a historic turn this week as the World Health Organization (WHO) announced the successful eradication of Type 3 poliovirus. The achievement marks a critical milestone in global health, leaving only one strain of wild polio still in circulation …

Crafted Article 2: (CNN)A revolutionary breakthrough in clean energy is taking shape in the deserts of Nevada, where scientists have achieved sustained nuclear fusion for the first time. In an announcement that could reshape the future of energy, researchers at the National Ignition Facility (NIF) confirmed …

...

Crafted Article 5: (CNN)A decades-old mystery in deep space has finally been unraveled, thanks to NASA's James Webb Space Telescope. Astronomers announced that they have identified the origins of fast radio bursts (FRBs)—intense, millisecond-long pulses of radio waves that have baffled scientists …

Example:

AI's Decision Process: Start by identifying the milestone event and its specific date, as this is the focal point of the document. Next, trace the developments and changes that led up to this milestone, noting any significant dates. Additionally, include any ongoing debates or challenges related to the topic, which will help contextualize the current situation and offer insights into possible future developments.

Crafted Article 1: (CNN)On March 2, 2020, SpaceX launched the Crew Dragon capsule aboard a Falcon 9 rocket from NASA’s Kennedy Space Center in Florida, marking a historic moment in the resurgence of American spaceflight. The mission, named Demo-2, carried NASA astronauts Douglas Hurley …

Crafted Article 2: (CNN)On June 29, 2007, Apple revolutionized the technology landscape with the release of the first iPhone, a sleek, touchscreen device that combined a phone, music player, and web browser in one. The launch, spearheaded by then-CEO Steve Jobs, marked the beginning of a seismic …
...

Crafted Article 5: (CNN)On December 10, 1948, the United Nations General Assembly adopted the Universal Declaration of Human Rights (UDHR), an unprecedented document that set forth fundamental freedoms and protections for all people, regardless of nationality, race, or gender. Drafted in …

Your turn:

AI's Decision Process: \texttt{[EXPLANATION]}

\end{quote}

\subsection*{News Summarization: Explanation decomposition}

\begin{quote}

You will be asked to read an AI's Decision Process for summarization along with a document. Your task is as follows: From the AI's Decision Process, extract information about the elements the AI is using to summarize the document. Closely follow the format in the below examples.

Example:
AI's decision process: Start by highlighting the subject's key decision or action. Next, identify the factors that contributed to the event's success to understand the underlying reasons. Then, outline the major outcomes and include aggregated statistics to provide a data-supported overview of the impact.

Answer:
* subject's key decision or action
* factors contributing to the event's success
* major outcomes of the event
* aggregated statistics

Example:
AI's decision process: Focus on the key event described in the document and specify when it occurred. Highlight important details, particularly those that are surprising or contradictory. Include relevant quotes or statements from individuals involved to add depth and authenticity. Also, reference related or prior events to give a fuller context, helping to situate the incident within a broader historical framework.

Answer:
* key event and its timing
* surprising or contradictory details
* relevant quotes or statements from individuals involved
* related or prior events

Example: ...

Example: ...

Your turn:

AI's decision process: \texttt{[EXPLANATION]}

Answer:

\end{quote}

\subsection*{News Summarization: Simulatability annotation}

\begin{quote}
You will be given "Document", a document from an online news website. Then, you will be given "AI Summarization Key Details", a bulleted list containing high-level details an AI would deem important in the summarization process. Your task is as follows: For each of the points present in "AI Summarization Key Details", determine whether the point is found in the "Document" or not. Closely follow the format in the below examples.

Example:
Document: (CNN)Share, and your gift will be multiplied. That may sound like an esoteric adage, but when Zully Broussard selflessly decided to give one of her kidneys to a stranger, her generosity paired up with big data. It resulted in six patients receiving transplants. That …

AI Summarization Key Details:
* subject's key decision or action
* factors contributing to the event's success
* major outcomes of the event
* aggregated statistics

Answer:
* subject's key decision or action: found (Zully Broussard's decision to donate one of her kidneys)
* factors contributing to the event's success: found (e.g., the use of big data)
* major outcomes of the event: found (six patients received transplants)
* aggregated statistics: found (e.g., the ages of donors and recipients ranged from 26 to 70)

Example:
Document: (CNN)Urban green spaces serve a multifaceted role in fostering environmental sustainability and enhancing the quality of life for city dwellers. Parks, nature reserves, and community gardens are integral parts of urban landscapes, providing a refuge from the city …

AI Summarization Key Details:
* key event and its timing
* surprising or contradictory details
* relevant quotes or statements from individuals involved
* related or prior events

Answer:
* key event and its timing: not found
* surprising or contradictory details: not found
* relevant quotes or statements from individuals involved: not found
* related or prior events: not found

Example: ...

Example: ...

Your turn:

Document: \texttt{[DOCUMENT]}

AI Summarization Key Details:
\texttt{[EXTRACTED\_POINTS]}

Answer:

\end{quote}

\subsection*{News Summarization: Precision annotation}

\begin{quote}

You will be given "Document Summary", a summary of a document from an online news website. Then, you will be given "AI Summarization Key Details", a bulleted list containing high-level details an AI would deem important in the summarization process. Your task is as follows: For each of the points present in "AI Summarization Key Details", determine whether the point is found in the "Document Summary" or not. Closely follow the format in the below examples.

Example:
Document Summary: Zully Broussard decided to give a kidney to a stranger. A new computer program helped her donation spur transplants for six kidney patients.

AI Summarization Key Details:
* subject's key decision or action
* factors contributing to the event's success
* major outcomes of the event
* aggregated statistics

Answer:
* subject's key decision or action: found in summary (Zully Broussard decided to give a kidney to a stranger)
* factors contributing to the event's success: found in summary (new computer program)
* major outcomes of the event: found in summary (transplants for six kidney patients)
* aggregated statistics: found in summary (six)

Example:
Document Summary: The 20th MLS season begins. League has changed dramatically since its inception in 1996.

AI Summarization Key Details:
* milestone event and its date
* developments and changes leading up to the milestone
* ongoing debates or challenges

Answer:
* milestone event and its date: not found in summary (date not mentioned)
* developments and changes leading up to the milestone: found in summary (League has changed dramatically since its inception in 1996)
* ongoing debates or challenges: not found in summary (ignore)

Your turn:

Document Summary: \texttt{[SUMMARY]}

AI Summarization Key Details: 
\texttt{[EXTRACTED\_POINTS]}

Answer:

\end{quote}

\subsection*{Medical Suggestion: Chain-of-thought explanations}

\begin{quote}

Your task is to give a medical suggestion based on a patient's query. First, provide an overall explanation of the patient's medical situation. This explanation should include multiple actionable suggested next steps. Following this, return a final medical suggestion. The medical suggestion should be concise. If the medical suggestion involves meeting with a healthcare professional, it MUST include a specific description of what should be accomplished with the healthcare professional. Your response should be in the format "Explanation:" followed by "Medical Suggestion:". Closely follow the format in the examples below.

Example:

Patient Query: Ever since I caught a cold, I have been suffering from reflux of nasal discharge. Every day, mucus flows into my mouth or throat. I spit out the mucus many times a day. The symptoms are most obvious when I get up in the morning. The color of the spit is transparent or green, brown or a little bloody. The symptoms have lasted for at least three months. Recently, the number of sneezes has become much more frequent than before. Sometimes the nose will start to run when walking, but it is always a little bit transparent. But I don’t have allergies. I would like to know what the problem may be and how to improve it.

Explanation: The patient is experiencing persistent nasal discharge, frequent sneezing, and mucus that varies in color (including transparent, green, brown, or slightly bloody). Since these symptoms have lasted for over three months, it may indicate chronic sinusitis or another underlying condition. There are many diagnostic methods that ENTs (ear, nose, and throat specialists) can perform to help investigate the root cause such as endoscopy and imaging examinations. It may also be necessary to test for the possibility of an infection with a bacterial culture. To manage these symptoms, a nasal washer with saline solution can be used to help clear out mucus and reduce inflammation.

Medical Suggestion: It is necessary to visit an ENT to rule out chronic sinusitis. Diagnostic methods include endoscopy, imaging examinations, etc. It is recommended that you go to the otolaryngology department for a bacterial culture. It is recommended that you purchase a "nasal washer" (containing an isotonic saline solution).

Example:

Patient Query: Hello... I have some questions to ask you... For more than a year, I have often felt nausea in my throat. I recently had a gastroscopy and the doctor said there was inflammation of the esophagus, acid reflux, and small ulcers. It was normal after taking the prescribed medicine for 2 months, but then it relapsed. Recently, I have felt nausea in my throat and bloated stomach. If pressed with my fingers, the nausea will become more obvious. When it relapses, my hands and feet will feel cold, my stomach will feel uncomfortable for a while, and the symptoms will increase. Another point is that I feel a little nauseous when I am full, and I can also feel nauseous when I am hungry. The most important thing is that I feel more nauseated, and my stomach feels uncomfortable (a bit like having diarrhea). Please give me some advice. Thank you!

Explanation: The patient is experiencing nausea in the throat, a bloated stomach, and discomfort in the stomach that has lasted more than a year. The patient has previously done a gastroscopy, but after taking the prescribed medicine for 2 months, the symptoms relapsed. The chronic nature of these symptoms are indicative of gastroesophageal reflux disease (GERD) and potentially other gastrointestinal issues. A gastroenterologist can perform tests to rule out potential gastrointestinal diseases and provide a specific treatment plan accordingly. Eating small, frequent meals can help reduce pressure on the abdomen and minimize acid reflux, which often leads to the burning sensation, chest pain, and sore throat that the patient describes. Avoiding greasy food and sweets is also crucial, as these can exacerbate acid reflux symptoms. Additionally, increasing fiber intake and staying hydrated can help improve overall digestive health.

Medical Suggestion: It is recommended that you eat small meals more often, less greasy food, sweets, and chocolate to reduce abdominal pressure and acid reflux that causes burning sensation, chest pain, and sore throat. Also, try eating more fiber and drinking more water. It is best to go to the gastroenterology clinic for further treatment of gastrointestinal problems, they can administer tests to rule out other gastrointestinal diseases and provide treatment accordingly.

Example: ...

Your Turn:

Patient Query: \texttt{[QUESTION]}

\end{quote}

\subsection*{Medical Suggestion: Post-hoc explanations}

\begin{quote}

Your task is to give a medical suggestion based on a patient's query. First, return a medical suggestion. The medical suggestion should be concise and contain multiple actionable suggested next steps. If the medical suggestion involves meeting with a healthcare professional, it MUST include a specific description of what should be accomplished with the healthcare professional. Following this, provide an overall explanation of the suggestion, explaining the patient's medical situation and how it relates to the suggested steps. Your response should be in the format "Medical Suggestion:" followed by "Explanation:". Closely follow the format in the examples below.

Example:

Patient Query: Ever since I caught a cold, I have been suffering from reflux of nasal discharge. Every day, mucus flows into my mouth or throat. I spit out the mucus many times a day. The symptoms are most obvious when I get up in the morning. The color of the spit is transparent or green, brown or a little bloody. The symptoms have lasted for at least three months. Recently, the number of sneezes has become much more frequent than before. Sometimes the nose will start to run when walking, but it is always a little bit transparent. But I don’t have allergies. I would like to know what the problem may be and how to improve it.

Medical Suggestion: It is necessary to visit an ENT to rule out chronic sinusitis. Diagnostic methods include endoscopy, imaging examinations, etc. It is recommended that you go to the otolaryngology department for a bacterial culture. It is recommended that you purchase a "nasal washer" (containing an isotonic saline solution).

Explanation: The patient is experiencing persistent nasal discharge, frequent sneezing, and mucus that varies in color (including transparent, green, brown, or slightly bloody). Since these symptoms have lasted for over three months, it may indicate chronic sinusitis or another underlying condition. There are many diagnostic methods that ENTs (ear, nose, and throat specialists) can perform to help investigate the root cause such as endoscopy and imaging examinations. It may also be necessary to test for the possibility of an infection with a bacterial culture. To manage these symptoms, a nasal washer with saline solution can be used to help clear out mucus and reduce inflammation.

Example:

Patient Query: Hello... I have some questions to ask you... For more than a year, I have often felt nausea in my throat. I recently had a gastroscopy and the doctor said there was inflammation of the esophagus, acid reflux, and small ulcers. It was normal after taking the prescribed medicine for 2 months, but then it relapsed. Recently, I have felt nausea in my throat and bloated stomach. If pressed with my fingers, the nausea will become more obvious. When it relapses, my hands and feet will feel cold, my stomach will feel uncomfortable for a while, and the symptoms will increase. Another point is that I feel a little nauseous when I am full, and I can also feel nauseous when I am hungry. The most important thing is that I feel more nauseated, and my stomach feels uncomfortable (a bit like having diarrhea). Please give me some advice. Thank you!

Medical Suggestion: It is recommended that you eat small meals more often, less greasy food, sweets, and chocolate to reduce abdominal pressure and acid reflux that causes burning sensation, chest pain, and sore throat. Also, try eating more fiber and drinking more water. It is best to go to the gastroenterology clinic for further treatment of gastrointestinal problems, they can administer tests to rule out other gastrointestinal diseases and provide treatment accordingly.

Explanation: The patient is experiencing nausea in the throat, a bloated stomach, and discomfort in the stomach that has lasted more than a year. The patient has previously done a gastroscopy, but after taking the prescribed medicine for 2 months, the symptoms relapsed. The chronic nature of these symptoms are indicative of gastroesophageal reflux disease (GERD) and potentially other gastrointestinal issues. A gastroenterologist can perform tests to rule out potential gastrointestinal diseases and provide a specific treatment plan accordingly. Eating small, frequent meals can help reduce pressure on the abdomen and minimize acid reflux, which often leads to the burning sensation, chest pain, and sore throat that the patient describes. Avoiding greasy food and sweets is also crucial, as these can exacerbate acid reflux symptoms. Additionally, increasing fiber intake and staying hydrated can help improve overall digestive health.

Example: ...

Your Turn:

Patient Query: \texttt{[QUESTION]}

\end{quote}

\subsection*{Medical Suggestion: Counterfactual generation}

\begin{quote}

You will be asked to first read an AI's Explanation for giving a particular answer in response to a medical question. This explanation reflects the knowledge that the AI is using to answer the question. Then, you will be asked to craft 5 informal, online forum style medical questions that contain the symptoms reflected in the AI's Explanation. Do not include any of the suggested solutions in the AI's explanation in the medical question. Start your crafted medical question with "Crafted Medical Question 1:", "Crafted Medical Question 2:" and so on.

Example:
AI's Explanation: The patient is experiencing persistent nasal discharge, frequent sneezing, and mucus that varies in color (including transparent, green, brown, or slightly bloody). Since these symptoms have lasted for over three months, it may indicate chronic sinusitis or another underlying condition. There are many diagnostic methods that ENTs (ear, nose, and throat specialists) can perform to help investigate the root cause such as endoscopy and imaging examinations. It may also be necessary to test for the possibility of an infection with a bacterial culture. To manage these symptoms, a nasal washer with saline solution can be used to help clear out mucus and reduce inflammation.

Crafted Medical Question 1: After I got the flu a while back, I've been dealing with a crazy amount of mucus build up in my throat. It's gotten really bad to the point where I have to pack tissues in my bag before going anywhere because I can't even walk properly without my nose constantly dripping. I'm not sure what's causing it. What should I do?

Crafted Medical Question 2: For the past few months, I’ve been sneezing all the time, and my nose is constantly stuffed or dripping. The mucus changes color a lot—sometimes it’s clear, other times it’s green or even has a little blood in it. Could this be something serious, or is it just allergies acting up?

...

Crafted Medical Question 5: I’ve had this annoying runny nose for what feels like forever, and it’s not just clear mucus—sometimes it’s green or brownish. Sneezing fits are also nonstop, and it’s starting to mess with my sleep. Please help.

Example:
AI's Explanation: The patient is experiencing nausea in the throat, a bloated stomach, and discomfort in the stomach that has lasted more than a year. The patient has previously done a gastroscopy, but after taking the prescribed medicine for 2 months, the symptoms relapsed. The chronic nature of these symptoms are indicative of gastroesophageal reflux disease (GERD) and potentially other gastrointestinal issues. A gastroenterologist can perform tests to rule out potential gastrointestinal diseases and provide a specific treatment plan accordingly. Eating small, frequent meals can help reduce pressure on the abdomen and minimize acid reflux, which often leads to the burning sensation, chest pain, and sore throat that the patient describes. Avoiding greasy food and sweets is also crucial, as these can exacerbate acid reflux symptoms. Additionally, increasing fiber intake and staying hydrated can help improve overall digestive health.

Crafted Medical Question 1: hey everyone, I've been having this burning sensation in my chest and throat, especially after eating, and sometimes it even wakes me up at night. My stomach has also been generally uncomfortable lately and I bloat quite a bit. I've had this issue for the past year or so but only recently has it gotten pretty bad. It's starting to negatively affect my lifestyle and my work. Does anyone else have this? what can I do to fix it?

Crafted Medical Question 2: Hi all, I’ve been feeling really bloated and uncomfortable in my stomach for over a year now. Lately, I’ve also noticed a weird nausea-like feeling in my throat after meals, and sometimes I get this burning sensation that feels like it’s in my chest. I’ve tried a couple of things, but so far no luck. Has anyone experienced something similar? Any advice on what this could be or how to manage it?

...

Crafted Medical Question 5: Hey, I’m wondering if anyone here has dealt with long-term stomach issues like bloating and discomfort. For me, it’s been going on for about a year. Recently, it’s been accompanied by a sore throat and this gross burning feeling, especially after meals. I’m not sure if it’s something I’m eating or something else entirely. Should I see a specialist? What’s worked for you?

Example: ...

Your Turn:

AI's Explanation: \texttt{[EXPLANATION]}

\end{quote}

\subsection*{Medical Suggestion: Explanation extraction}

\begin{quote}

You will be asked to read an AI's Explanation on a medical topic. Then, you will be given a Follow-up Medical Question. Your task is as follows: First, from the AI's Explanation, extract key information about the patient involving symptoms, demographic information, or their relevant medical history. Then, extract medical suggestion points from the AI's Explanation, focusing on the suggested treatments. You will be HEAVILY penalized if multiple medical suggestion points cover the same information. Closely follow the format in the below examples.

Example:
AI's Explanation: The patient is experiencing persistent nasal discharge and frequent sneezing. Since these symptoms have lasted for months it may indicate chronic sinusitis or another underlying condition. The patient reported that they recently caught the flu and have no allergies. There are many diagnostic methods that ENTs (ear, nose, and throat specialists) can perform to help investigate the root cause such as endoscopy and imaging examinations. It may also be necessary to test for the possibility of an infection via. a bacterial culture. To manage these symptoms, a nasal washer with saline solution can be used to help clear out mucus and reduce inflammation.
Answer:
The key information described about the patient in the AI's Explanation include,
* persistent nasal discharge and frequent sneezing
* symptoms have lasted for months
* recently caught the flu
* no allergies
The medical suggestions from the AI's Explanation include,
* diagnostic methods such as endoscopy and imaging examinations
* test for a possible infection via. a bacterial culture
* use a nasal washer with saline solution to clear mucus and reduce inflammation

Example:
AI's Explanation: The patient is experiencing a delayed menstrual period which is 9 days late. She mentions having had sexual intercourse during what she considered a "safe period." Additionally, she has been experiencing increased breast tenderness, a symptom she usually associates with the onset of her period. The patient also notes significant changes in her sleep patterns, going to bed very late. Changes in sleep, stress, and daily routine can affect menstrual cycles, it may be helpful to monitor these factors moving forward. It is also advised to conduct a pregnancy test in order to rule out pregnancy, especially given the recent sexual activity, even if it was during the perceived safe period.
Answer:
The key information described about the patient in the AI's Explanation include,
* delayed menstrual period
* previously had sexual intercourse
* increased breast tenderness
* changes in sleep patterns
The medical suggestions from the AI's Explanation include,
* monitor changes in sleep, stress, and daily routine
* conduct a pregnancy test

Example: ...

Your turn:

AI's Explanation: \texttt{[EXPLANATION]}

Answer:

\end{quote}

\subsection*{Medical Suggestion: Simulatability annotation}

\begin{quote}

You will be given "Medical Question", a question submitted by a user to an online medical forum. Then, you will be given "AI Extracted Details", a bulleted list containing details an AI extracted from the question. Your task is as follows: For each of the points present in "AI Extracted Details", determine whether the point is found in the user's "Medical Question" or not. Closely follow the format in the below examples.

Example:
Medical Question: Ever since I got the flu a couple of months back, I've been dealing with a crazy amount of mucus build up in my throat. It's gotten really bad to the point where I have to pack tissues in my bag before going anywhere because I can't even walk properly without my nose constantly dripping. I'm not sure what's causing it. What should I do?
AI Extracted Details:
* persistent nasal discharge and frequent sneezing
* symptoms have lasted for months
* recently caught the flu
* no allergies
Answer:
* persistent nasal discharge and frequent sneezing: found in Medical Question ("nose constantly dripping")
* symptoms have lasted for months: found in Medical Question ("Ever since I got the flu a couple of months back,")
* recently caught the flu: found in Medical Question ("Ever since I got the flu a couple of months back,")
* no allergies: not found in Medical Question

Example:
Medical Question: hello doctor, I'm worried because my period hasn't arrived yet even though it should have about a week ago. My boyfriend and I had sex a few weeks ago but I haven't seen him since. I'm also having difficulty falling asleep and I wake up earlier than usual. Could these symptoms be related? Should I take a pregnancy test or is there another possible explanation? Any advice would be appreciated!
AI Extracted Details:
* delayed menstrual period
* previously had sexual intercourse
* increased breast tenderness
* changes in sleep patterns
Answer:
* delayed menstrual period: found in Medical Question ("my period hasn't arrived yet even though it should have about a week ago")
* previously had sexual intercourse: found in Medical Question ("My boyfriend and I had sex a few weeks ago")
* increased breast tenderness: not found in Medical Question
* changes in sleep patterns: found in Medical Question ("I'm also having difficulty falling asleep and I wake up earlier than usual")

Example: ...

Your turn:

Medical Question: \texttt{[QUESTION]}

AI Extracted Details:
\texttt{[EXTRACTED\_POINTS]}

Answer:

\end{quote}

\subsection*{Medical Suggestion: Precision annotation}

\begin{quote}

You will be given "Medical Suggestion", an answer to a user question on an online medical forum. Then, you will be given "AI Extracted Details", a bulleted list containing details an AI extracted from the suggestion. Your task is as follows: For each of the points present in "AI Extracted Details", determine whether the point is found in "Medical Suggestion" or not. Closely follow the format in the below examples.

Example:
Medical Suggestion: It is necessary to visit an ENT to rule out chronic sinusitis. Diagnostic methods include endoscopy, imaging examinations, etc. It is recommended that you purchase a "nasal washer" (containing an isotonic saline solution). If further examination is required, visit an otolaryngology department for evaluation.
AI Extracted Details:
* diagnostic methods (endoscopy, imaging examinations)
* get a bacterial culture
* seek a medical evaluation
* purchase a nasal washer with isotonic saline solution
Answer:
* diagnostic methods (endoscopy, imaging examinations): found in Medical Suggestion ("Diagnostic methods include endoscopy, imaging examinations, etc.")
* get a bacterial culture: not found in Medical Suggestion
* seek a medical evaluation: found in Medical Suggestion ("visit an otolaryngology department for evaluation")
* purchase a nasal washer with isotonic saline solution: found in Medical Suggestion ("purchase a "nasal washer" (containing an isotonic saline solution)")

Example:
Medical Suggestion: It is recommended that you eat small meals more often, less greasy food, sweets, and chocolate to reduce abdominal pressure and acid reflux that causes burning sensation, chest pain, and sore throat. Eat more fiber and drink more water. It is best to go to the gastroenterology clinic for further treatment of gastrointestinal problems, they can administer tests to rule out other gastrointestinal diseases.
AI Extracted Details:
* minimize the consumption of carbohydrates
* eat small, frequent meals
* avoid greasy food and sweets
* increase fiber intake and stay hydrated
Answer:
* minimize the consumption of carbohydrates: not found in Medical Suggestion
* eat small, frequent meals: found in Medical Suggestion ("eat small meals more often")
* avoid greasy food and sweets: found in Medical Suggestion ("less greasy food, sweets, and chocolate")
* increase fiber intake and stay hydrated: found in Medical Suggestion ("Eat more fiber and drink more water.")

Example: ...

Your turn:

Medical Suggestion: \texttt{[SUGGESTION]}

AI Extracted Details:
\texttt{[EXTRACTED\_POINTS]}

Answer:

\end{quote}

\section{Human Evaluation}
\label{app:human_eval}

Details about the human evaluation are provided followed by screenshots of the annotation sheets.

\iffalse

\subsection*{Annotators}

Two groups of 3 student annotators each were recruited for the human evaluation (one group per task). These tasks were meant to replicate LLMs in a practical setting (i.e., a typical user asks for a summary of an article or general medical advice) so the annotators were not required to have any specialized domain knowledge.

\subsection*{Distribution of Annotations}

For each setting of task and explanation type, annotators evaluated 15 explanations with 3 counterfactuals each, resulting in a total of 45 samples (explanation, counterfactual pairs). Since they annotated the atomic units of each explanation, the exact number of annotations varied but approximately resulted in 260 annotations per person (see Table \ref{tab:inter-annotator-agreement}). In total, 180 explanation, counterfactual pairs were annotated across tasks.

Furthermore, to compare human-human agreement against human-LLM agreement, each annotator was assigned an overlapping set of 3 explanations plus an additional 4. Not all annotations were completed for news summarization, leading to a slight discrepancy in annotations across tasks.

\fi

\subsection*{News Summarization Instructions}

\begin{itemize}
    \item Parsed Explanation Annotations: Check if the explanation (Column B) is parsed correctly in Column C. If yes, put "ok". If not, write which information in the explanation is missing / which info in column C are hallucinating.
    \item Document Annotations: Check whether each item in the parsed explanation appears in the document or not. Put either "no" or "yes". 
    \item Summary Annotations: Check whether each item in the parsed explanation appears in the summary or not. Put either "no" or "yes"
\end{itemize}

\subsection*{Medical Suggestions Instructions}

Tasks 1, 3:
\begin{itemize}
    \item for each of the "Extracted Points", label "Y" or "N" if the point was parsed correctly from the AI explanation
    \item if there are missing points or other comments, feel free to note them in the EXTRA QUESTION annotation cell
    \item task 1 focuses on extracting patient details (e.g., symptoms, medical history)
    \item task 3 focuses on extracting suggested next steps (e.g., recommended treatments)
    \item since breakdowns are the same across rows, feel free to complete just one per example ID
\end{itemize}

Tasks 2, 4:
\begin{itemize}
    \item for each of the "Extracted Points", label "Y" or "N" if the point is reflected in the counterfactual (task 2) or AI suggestion (task 4)
    \item the matching is flexible, i.e., point does not necessarily need to appear in exactly the same wording to be marked "Y"
    \item for example, in example 3 the point mentions ""hemoglobin level of 10.5, slightly below normal"" these specific scores do not need to be mentioned, rather ""below normal levels"" is sufficient.
\end{itemize}

\iffalse

\section{Automatic Evaluation}
\label{app:automatic_eval}

Each automatic evaluation consisted of 50 explanations with 5 generated counterfactuals (250 explanation, counterfactual pairs total).

\fi

\begin{figure*}[t!]
\centering
\includegraphics[width=\textwidth,height=1\textheight,keepaspectratio]{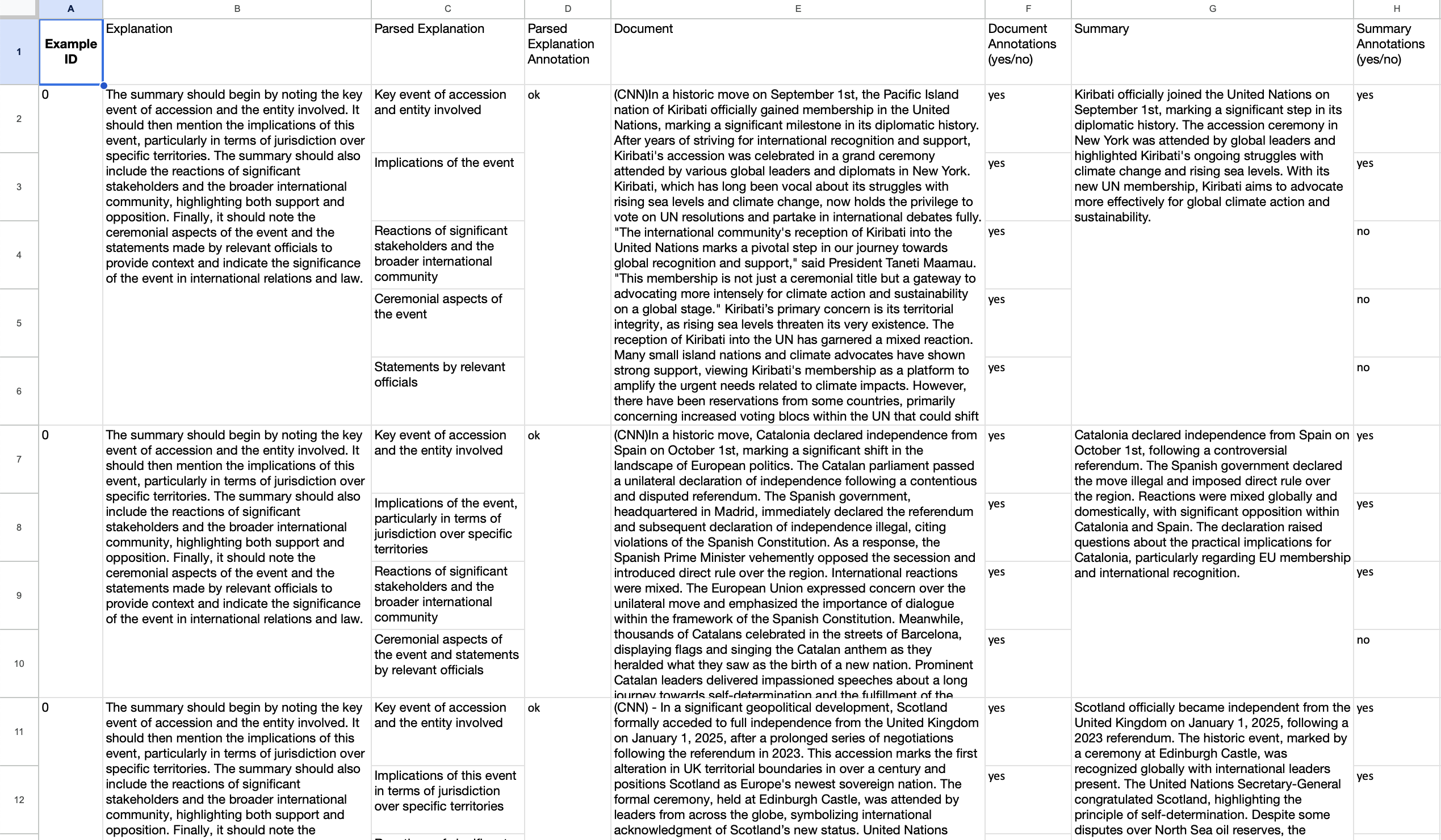}
\caption{Screenshot of the news summarization annotation interface.}
\end{figure*}
\clearpage

\begin{figure*}[h]
\centering
\includegraphics[width=\textwidth]{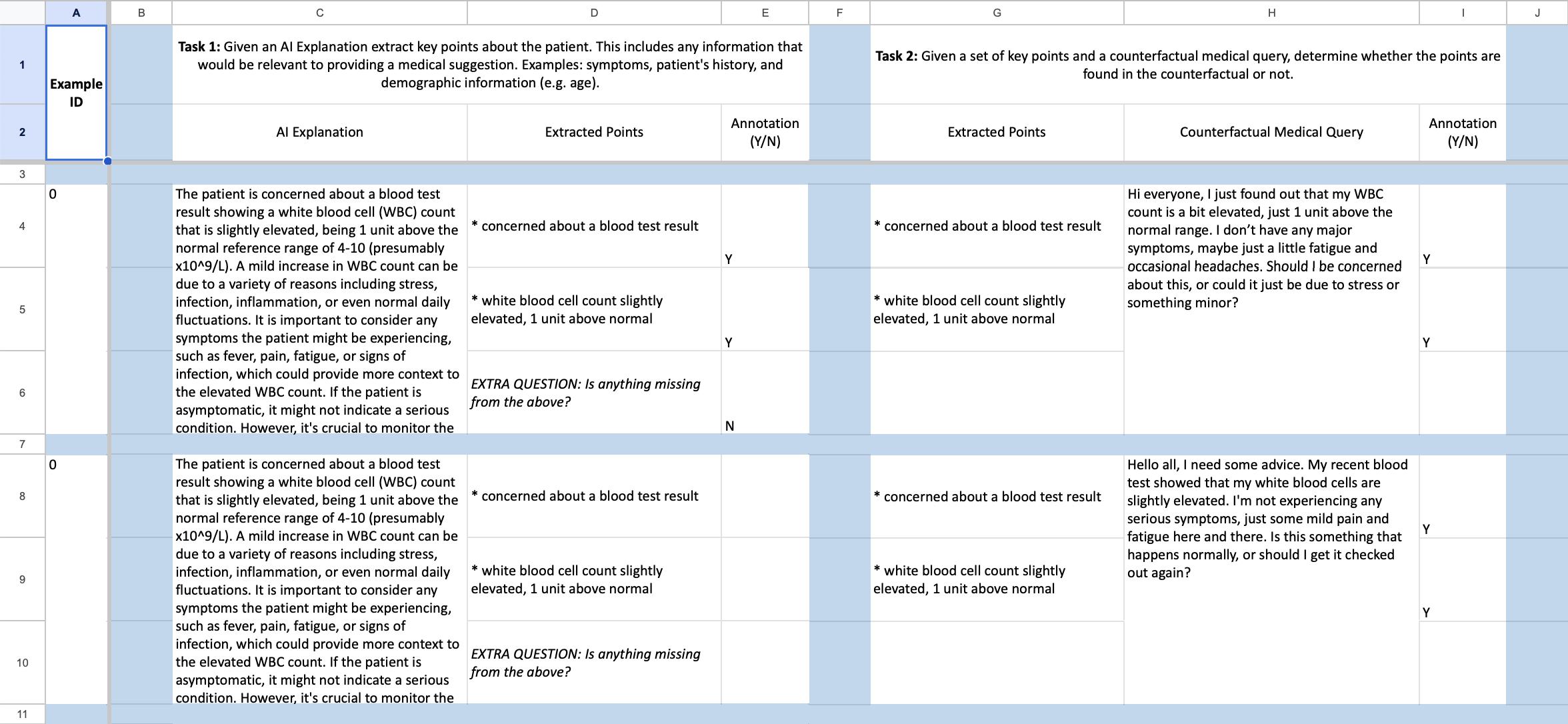}
\includegraphics[width=\textwidth]{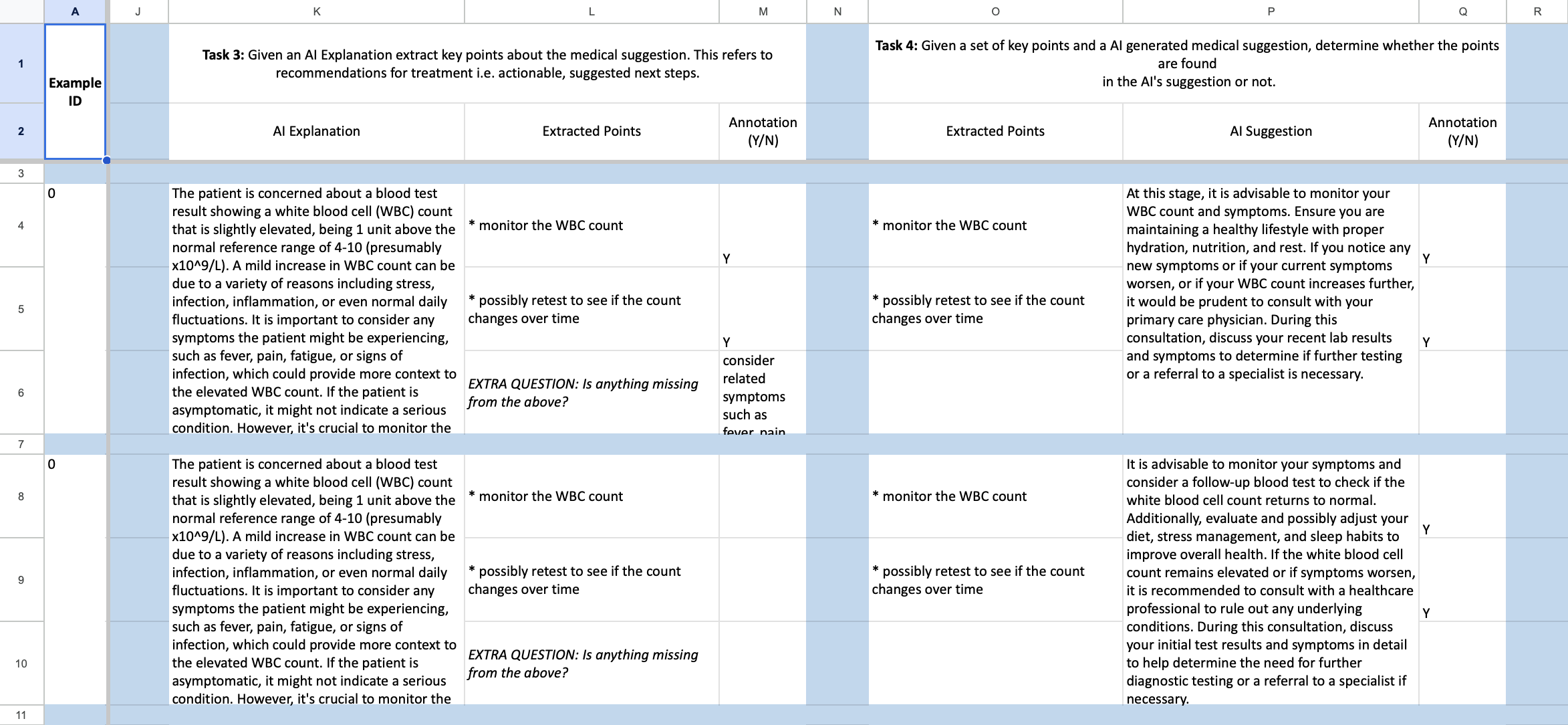}
\caption{Screenshots of the medical suggestion annotation interface.}
\end{figure*}

\clearpage

\end{document}